\documentclass[preprint,twoside,english,sort&compress]{elsarticle}

\usepackage[a4paper,left=30mm,width=150mm,top=25mm,bottom=25mm]{geometry}
\usepackage[T1]{fontenc}
\usepackage[latin9]{inputenc}
\usepackage{amstext}
\makeatletter
\journal{}
\usepackage{xcolor}
\usepackage{amsmath}
\usepackage{amssymb}
\usepackage{slashbox,pict2e} 
\usepackage{ctex}

\usepackage{Sweave}

\newtheorem{lemma}{Lemma}

\newtheorem{definition}{Definition}

\makeatother
\usepackage{amsmath,bm}
\usepackage{algorithm}
\usepackage{algorithmic}
\usepackage{babel}
\usepackage{setspace}
\usepackage[colorlinks,linkcolor=red]{hyperref}
\def\varneg{\mathord{\raise1.41ex\hbox{\vrule width.4em height.4pt}\kern0pt\vrule width.4pt height1.5ex}}
\graphicspath{{./fig/}}
\begin{document}

\begin{frontmatter}

\title{Median evidential $c$-means algorithm and its application to community detection}


\author[rvt,focal]{Kuang Zhou\corref{cor1}}

\ead{kzhoumath@163.com}

\author[focal]{Arnaud Martin}

\ead{Arnaud.Martin@univ-rennes1.fr}

\author[rvt]{Quan Pan} \author[rvt]{Zhun-ga Liu}
\cortext[cor1]{Corresponding author}
\address[rvt]{School of Automation, Northwestern Polytechnical University,
Xi'an, Shaanxi 710072, PR China} \address[focal]{IRISA, University of Rennes
1, Rue E. Branly, 22300 Lannion, France}

\begin{abstract}
Median clustering is of great value for partitioning relational data. In this paper, a new prototype-based    clustering method, called Median Evidential $C$-Means (MECM), which is an extension of  median $c$-means and  median fuzzy $c$-means on the theoretical framework of belief functions  is proposed. The median  variant relaxes the restriction of a metric space embedding for the objects but constrains the prototypes to be in the original data set. Due to    these properties, MECM could be applied to graph clustering problems. A community detection scheme for social   networks based on MECM is investigated and the obtained credal partitions of graphs, which are more refined than  crisp and fuzzy ones, enable us to have a better understanding of the graph structures. An initial prototype-selection  scheme based on evidential semi-centrality is presented to avoid local premature convergence and an evidential modularity function is defined to choose the optimal number of communities.  Finally,  experiments in  synthetic and   real data sets illustrate the performance of MECM and show its difference to other methods.
\end{abstract}


\end{frontmatter}

\section{Introduction}
 \renewcommand{\thefootnote}{}
\footnotetext{The figures are not displayed in this version due to the limitation of size. The complete version could be found in \url{http://www.sciencedirect.com/science/article/pii/S095070511400402X} or contact the authors to get the paper.}
Cluster analysis or clustering is the task of
partitioning a set of $n$ objects $X=\{\bm{x}_1,\bm{x}_2,\cdots,\bm{x}_n\}$
into $c$ small groups $\Omega=\{\omega_1,\omega_2,\cdots,\omega_c\}$ in such a
way that objects in the same group (called a cluster) are more similar (in
some sense or another, like characteristics or behavior) to each other than to
those in other groups. 
The clustering can be used in many fields such as privacy preserving \citep{islam2011privacy} , information retrieval \citep{bordogna2012quality}, text analysis \citep{zhang2010text},  etc. It can also
be used as the first step of classification problems  to identify the distribution of the training set \citep{yang2011kernel}.
Among the existing approaches to clustering, the objective function-driven or
prototype-based clustering  such as $c$-means and Gaussian mixture modeling is
one of the most widely applied paradigms in statistical pattern recognition.
These methods are based on a fundamentally very simple, but nevertheless very
effective idea, namely to describe the data under consideration by a set of
prototypes. They  capture the characteristics of the data distribution (like
location, size, and shape), and classify the data set based on
the similarities (or dissimilarities) of the objects to their prototypes
\citep{borgelt2006prototype}.

Generally, a $c$-partition of $n$ objects in $X$ is a set of $n\times c$
values $\{u_{ij}\}$  arrayed as an $n\times c$ matrix $\bm{U}$. Each element
$u_{ij}$ is the membership of $\bm{x}_i$ to cluster $j$. The classical $C$-Means (CM) method aims to
partition $n$  observations into $c$ groups in which each observation
belongs to the class with the nearest mean, serving as a prototype of the
cluster. It results in $u_{ij}$ is either 0 or 1 depending whether object
$i$ is  grouped into cluster $j$, and thus each data point is assigned to a
single cluster (hard partitions). Fuzzy $C$-Means (FCM),  proposed  by
\citet{dunn1973fuzzy}   and  later  improved  by  \citet{bezdek1981pattern},
is  an  extension  of  $c$-means  where  each  data  point  can  be  a  member
of  multiple clusters with membership values (fuzzy partitions)
\citep{jain2010data}.

Belief functions have already been used to express partial information about data both in supervised and unsupervised learning \citep{masson2008ecm,tabassian2012combining}. Recently, \citet{masson2008ecm} proposed the application of evidential $c$-means (ECM) to get credal partitions \citep{denoeux2004evclus} for object data. The credal partition is  a
general extension of the  crisp (hard) and fuzzy ones and it allows the object
not only to belong to single clusters, but also to belong to any subsets of
$\Omega$ by allocating a mass of belief for each object in $X$ over the power
set $2^{\Omega}$. The additional flexibility brought by the power set
provides more refined partitioning results than those by the other  techniques
allowing us to gain a deeper insight into the data \citep{masson2008ecm}.

All of these aforementioned partition approaches are prototype-based. In CM, FCM
and ECM, the prototypes of clusters are the geometric centers of included data
points in the corresponding groups. However, this may be inappropriate as it
is the case in community detection problems for social networks, where the
prototype (center) of one group is likely to be one of the persons ({\em i.e.}
nodes in the graph) playing the leader role in the community. That is to say,
one of the  points in the group is better to be selected as a prototype,
rather than the center of all the points. Thus we should set some  constraints
for the prototypes, for example, let them be data objects. Actually this
is the basic principle of median clustering methods
\citep{geweniger2010median}.  These restrictions on prototypes can relax
the assumption of a metric space embedding for the objects to be clustered \citep{geweniger2010median,hammer2007relational}, and only
similarity or dissimilarity between data objects is required. There are some
clustering methods for relational data,  such as Relational FCM (RFCM)
\citep{hathaway1989relational} and Relational ECM (RECM)
\citep{masson2009recm}, but an underlying metric
is assumed for the given dissimilarities between objects. However, in
median clustering this restriction is dropped \citep{geweniger2010median}.
\citet{cottrell2006batch} proposed Median $C$-Means clustering method (MCM)
which is a variant of the classic $c$-means and proved the convergence of the
algorithm. \citet{geweniger2010median} combined MCM with the fuzzy $c$-means
approach and investigated the behavior of the resulted Median Fuzzy $C$-means
(MFCM) algorithm.

Community detection, which can extract specific structures from complex
networks,  has attracted considerable attention crossing many areas
from physics, biology, and economics to sociology. 
Recently, significant progress has been achieved in this research field and several popular algorithms for community detection have been presented. 
One of the most popular type of classical methods partitions networks by optimizing some criteria.  \citet{newman2004finding} proposed a network modularity measure (usually denoted by $Q$) and several algorithms that try to maximize $Q$ have been
designed \citep{blondel2008fast,clauset2004finding,duch2005community,tasgin2007community}.  But recent researches have found that the modularity based algorithms could not detect communities smaller than a certain size.
This problem is famously known as the resolution limit \citep{fortunato2007resolution}.  The single optimization criteria {\em i.e.} modularity may not be adequate to represent the structures in complex networks, thus \citet{amiri2013community} suggested a new  community detection process as a multi-objective optimization problem.
Another family of approaches considers hierarchical clustering techniques. It merges or splits clusters according to a topological measure of similarity between the nodes and tries to  build a hierarchical tree of partitions \citep{yang2013hierarchical,lancichinetti2009detecting,sales2007extracting}.
Also there are some ways, such as spectral methods
\citep{smyth2005spectral} and signal process method
\citep{hu2008community,jiang2012efficient}, to map topological
relationship of nodes on networks into geometrical structures of vectors in
$n$-dimensional Euclidian space, where  classical clustering methods like
CM, FCM and ECM could be evoked. However, there must be some loss of accuracy after
the mapping process. As mentioned before, for community detection, the
prototypes should be some nodes in the graph. Besides, usually only
dissimilarities between nodes are known to us. Due to the application of the relaxation on the
data objects and the constraints on the prototypes, the median clustering
could be applied to the community detection problem in social networks.

In this paper, we extend the median clustering methods in the framework of
belief functions theory and put forward the Median Evidential $C$-Means (MECM)
algorithm. Moreover, a community detection scheme based on MECM is also presented. Here, we emphasize two
key points different from those earlier studies. Firstly,
the proposed approach could provide credal partitions for
data set with only known dissimilarities. The dissimilarity measure could be
neither symmetric nor fulfilling any metric requirements. It is only required to be of  intuitive meaning. Thus it expands application
scope of credal partitions.  Secondly, some practical issues about how to apply the
method into community detection problems such as how to determine the initial prototypes and the optimum community number in the sense of
credal partitions are discussed.  This makes the approach appropriate for graph partitions and gives us a better understanding of the analysed
networks, especially for the uncertain and imprecise structures.

The rest of this paper is organized as follows:  Section 2 recalls the necessary
background related to this paper. In Section 3, the median $c$-means algorithm
is presented and in section 4, we show how the proposed method could be
applied in  the  community detection problem. In order to show the
effectiveness of our approach, in section 5 we test our algorithm on
artificial and real-world data sets  and make comparisons with different
methods. The final section makes the conclusions.

\section{Background}
\subsection{Theory of belief functions}
Let $\Omega=\{\omega_{1},\omega_{2},\ldots,\omega_{c}\}$ be the finite domain of
$X$, called the discernment frame. The mass function is defined on the power
set $2^{\Omega}=\{A:A\subseteq\Omega\}$.

\begin{definition}
The function $m:2^{\Omega}\rightarrow[0,1]$ is said to be the Basic Belief
Assignment (bba) on $\text{2}^{\Omega}$, if it satisfies:
\begin{equation}
\sum_{A\subseteq\Omega}m(A)=1.
\end{equation}
Every $A\in2^{\Omega}$ such that $m(A)>0$ is called a focal element.  The
credibility and plausibility functions are defined in Eq.$~\eqref{bel}$ and
Eq.$~\eqref{pl}$.
 \end{definition}
\begin{equation}
Bel\text{(}A\text{)}=\sum_{B\subseteq A, B \neq \emptyset} m\text{(}B\text{)},\forall A\subseteq\Omega,
\label{bel}
\end{equation}
\begin{equation}
 Pl\text{(}A\text{)}=\sum_{B\cap A\neq\emptyset}m\text{(}B\text{)},\forall A\subseteq\Omega.
 \label{pl}
\end{equation}
Each quantity $Bel(A)$  measures the total support given to $A$, while $Pl(A)$ can be
interpreted as the degree to which the evidence fails to
support the complement of $A$. The function
$pl:\Omega\rightarrow[0,1]$  such that $pl(\omega_i)=Pl(\{\omega_i\})$
$(\omega_i \in \Omega)$ is called the contour function associated with $m$.
A belief function on the credal level can be transformed into a probability function by Smets method. In this
algorithm, each mass of belief $m(A)$ is
equally distributed among the elements of $A$~\citep{smets2005decision}.
This leads to the concept of pignistic probability, $BetP$, defined by
\begin{equation}
 \label{pig}
	BetP(\omega_i)=\sum_{\omega_i  \in A \subseteq \Omega } \frac{m(A)}{|A|(1-m(\emptyset))},
\end{equation}
where $|A|$ is the number of elements of $\Omega$ in $A$.

Pignistic probabilities, which play the same role as fuzzy membership, can
easily help us  make a decision. In fact, belief functions provide us many
decision-making techniques not only  in the form of probability measures. For
instance, a pessimistic decision can be made by maximizing the credibility
function, while maximizing the plausibility function could provide an
optimistic one \citep{martin2008decision}. Another criterion
\citep{martin2008decision} considers the plausibility functions
and consists in attributing the class $A_j$ for  object $i$ if
\begin{equation}
\label{mbx1}
 A_j=\arg \max_{X \subseteq \Omega}
	\{m_i^b(X)Pl_i(X)\},
\end{equation}
where
\begin{equation} \label{mbx}
	m_i^b(X)=K_i^b \lambda_X \left(\frac{1}{|X|^r}\right).
\end{equation}
In Eq.~\eqref{mbx1} $m_i^b(X)$ is a weight on $Pl_i(X)$, and $r$ is a parameter
in $[0,1]$ allowing a decision from a simple class $(r= 1)$ until the total
ignorance $\Omega$ $(r= 0)$. The value $\lambda_X$ allows the integration of
the lack of knowledge on one of the focal sets $X\subseteq \Omega$, and it can
be set to be 1 simply. Coefficient $K_i^b$ is the normalization factor to
constrain the mass to be in the closed world:
 \begin{equation}
	K_i^b=\frac{1}{1-m_i(\emptyset)}.
\end{equation}

\subsection{Median $c$-means and median fuzzy $c$-means}
 Median c-means  is a variant of the
traditional $c$-means method \citep{cottrell2006batch,geweniger2010median}.
We assume that $n$ ($p$-dimensional)  data objects
$\bm{x}_i=\{x_{i1},x_{i2},\cdots,x_{ip}\}$ $(i=1,2,\cdots,n)$ are given. The
object set is denoted by $X=\{\bm{x}_1,\bm{x}_2,\cdots,\bm{x}_n\}$. The
objective function of MCM is similar to that in CM:
\begin{equation}
\label{J_mcm}
J_\text{MCM}=\sum_{j=1}^c \sum_{i=1}^n u_{ij}d^2_{ij},
\end{equation}
where $c$ is the number of clusters. As MCM is
based on crisp partitions, $u_{ij}$ is either 0 or 1 depending whether
$\bm{x}_i$ is in cluster $j$. The value $d_{ij}$ is the dissimilarity between
$\bm{x}_i$ and the prototype vector $\bm{v}_j$ of cluster $j$
$(i=1,2,\cdots,n, j=1,2,\cdots,c)$, which is not assumed to be fulfilling any
metric properties but  should reflect the common sense of dissimilarity. Due
to these weak assumptions, data object $x_i$ itself may be a  general
choice and it does not have to live in a metric  space
\citep{geweniger2010median}. The main difference between MCM and CM is that
the prototypes of MCM are restricted to   the data objects. 

Median fuzzy $c$-means (MFCM) merges MCM and the standard fuzzy $c$-means
(FCM). As in MCM, it requires the knowledge of the dissimilarity between
data objects, and the prototypes are restricted to the objects themselves
\citep{geweniger2010median}. MFCM also performs a two-step iteration scheme to
minimize the cost function
\begin{equation}\label{J_MFCM}
	J_\text{MFCM}=\sum_{j=1}^c \sum_{i=1}^n u_{ij}^\beta d^2_{ij},
\end{equation}
subject to the constrains
\begin{equation}
\sum_{k=1}^c u_{ik}=1, \forall i \in \{1,2,\cdots,n\},
\end{equation} and
\begin{equation}
 \sum_{i=1}^n u_{ik}>0, \forall k \in \{1,2,\cdots,c\},
\end{equation}
where each number $u_{ik}\in [0,1]$ is interpreted as a degree
of membership of object $i$ to cluster $k$, and $\beta>1$ is a weighting
exponent that controls the fuzziness of the partition. Again, MFCM is
preformed by alternating update steps as for MCM:
\begin{itemize}
\item
		Assignment update:
\begin{equation}
		u_{ij}=\frac{d_{ij}^{-2/(\beta-1)}}{\sum_{k=1}^c
	d_{ik}^{-2/(\beta-1)}}.
\end{equation}
\item Prototype update: the new
	prototype of cluster $j$ is set to be $\bm{v}_{j}=\bm{x}_{l^*}$ with
	\begin{equation} \bm{x}_{l^*}= \arg \min _{\{\bm{v}_j:\bm{v}_j=\bm{x}_l
	(\in X)\}} \sum_{i=1}^n u_{ij}^\beta d_{ij}^2.
 \end{equation}
\end{itemize}

\subsection{Evidential $c$-means}
Evidential $c$-means \citep{masson2008ecm} is a direct generalization of FCM in the
framework of belief functions, and it is based on the credal partition first
proposed by \citet{denoeux2004evclus}. The credal partition takes advantage
of imprecise (meta) classes \citep{liu2013evidential} to express partial knowledge of class memberships. The principle is different from
another belief clustering method  put forward by
\citet{schubert2004clustering}, in which  conflict
between evidence is utilised to cluster the belief functions related to
multiple events. In ECM, the evidential membership of an object
$\bm{x}_i=\{\bm{x}_{i1},\bm{x}_{i2},\cdots,\bm{x}_{ip}\}$ is represented by a
bba $\bm{m}_i=\left(m_i(A_j): A_j \subseteq \Omega \right)$ over the given
frame of discernment $\Omega=\{\omega_1,\omega_2,\cdots,\omega_c\}$. The
optimal credal partition is obtained by  minimizing the following objective
function:
\begin{equation}
	J_{\mathrm{ECM}}=\sum\limits_{i=1}^{n}\sum\limits_{A_j\subseteq \Omega,A_j
	\neq \emptyset}|A_j|^\alpha
	m_{i}(A_j)^{\beta}d_{ij}^2+\sum\limits_{i=1}^{n}\delta^2m_{i}(\emptyset)^{\beta}
	\label{JECM}
\end{equation}
constrained on
\begin{equation}
	\sum\limits_{A_j\subseteq \Omega,A_j \neq
	\emptyset}m_{i}(A_j)+m_{i}(\emptyset)=1, \label{ECMconstraint}
\end{equation}
where $m_{i}(A_j)\triangleq m_{ij}$ is the bba of $x_i$ given
to the nonempty set $A_j$, while $m_{i}(\emptyset)\triangleq m_{i\emptyset}$
is the bba of $x_i$ assigned to the emptyset, and $|\cdot|$ is the cardinality of
the set. Parameter $\alpha$ is a tuning parameter allowing to control the
degree of penalization for subsets with high cardinality, parameter $\beta$ is
a weighting exponent and  $\delta$ is an adjustable  threshold  for  detecting
the outliers. It is noted that for credal partitions, $j$ is not from 1 to $c$
as before, but ranges in $[0,f]$ with $f=2^c$. Here $d_{ij}$ denotes the
Euclidean distance between $x_i$ and the barycenter ({\em i.e.} prototype,
denoted by $\overline{\bm{v}}_j$) associated with $A_j$:
\begin{equation}\label{dis_node_pro}
  d_{ij}^2=\|x_i-\overline{\bm{v}}_j\|^2,
\end{equation}
where $\overline{\bm{v}}_j$ is defined mathematically by
\begin{equation}\label{prototypes}
	\overline{\bm{v}}_j=\frac{1}{|A_j|}\sum_{k=1}^c s_{kj} \bm{v}_k,
	~~\text{with}~~ s_{kj}=
  \begin{cases}
        1 & \text{if} ~~ \omega_k \in A_k \\
        0 & \text{else}\end{cases}.
 \end{equation}
The notation $\|\cdot\|$
denotes  the Euclidean norm of a vector, and $\bm{v}_k$ is the geometrical
center of all the points in cluster $k$. The update for ECM is given by the
following two alternating steps and the update formulas  can been obtained by
Lagrange multipliers method.
\begin{itemize}
\item Assignment update:
\begin{align}
\label{ECM_updata_bba}
m_{ij}=\frac{|A_j|^{-\alpha/(\beta-1)}{d_{ij}^{-2/(\beta-1)}}}{\sum\limits_{A_k\neq\emptyset}
|A_k|^{-\alpha/(\beta-1)}{d_{ik}^{-2/(\beta-1)}}+\delta^{-2/(\beta-1)}},
\forall i=1,2\cdots,n, \forall j/A_j \subseteq \Omega, A_j \neq \emptyset
\end{align}
 \begin{align}
			m_{i\emptyset}=1-\sum_{A_j\neq \emptyset}m_{ij}, \forall
		i=1,2,\cdots,n.
\end{align}
\item Prototype update: The prototypes
		(centers) of the classes are given by the rows of the matrix
		$v_{c\times p}$, which is the solution of the following linear system:
	\begin{equation}
    \bm{HV}=\bm{B},
   \end{equation}
where $\bm{H}$ is a matrix of size $(c \times c)$ given by
\begin{equation}\label{H}
   \bm{H}_{lk}=\sum_i
			\sum_{A_j \supseteqq \{\omega_k,\omega_l\}} |A_j|^{\alpha-2}
			m_{ij}^\beta, k,l=1,2,\cdots,c,
\end{equation} and $\bm{B}$ is a matrix of size $(c \times p)$ defined by
\begin{equation}\label{B}
			\bm{B}_{lq}=\sum_{i=1}^n x_{iq}\sum_{A_j\ni
			\omega_l}|A_j|^{\alpha-1}m_{ij}^\beta, l=1,2,\cdots,c,~~
			q=1,2,\cdots,p.
\end{equation} \end{itemize}
\subsection{Some concepts for social networks}
In this work we will investigate how the proposed clustering algorithm could be applied to community detection problems in social networks. In this section some concepts related to social networks will be recalled.
\subsubsection{Centrality and dissimilarity}
The problem of assigning centrality values to nodes  in graphs has been widely investigated as it is important for identifying influential nodes \citep{zhang2013identifying}. \citet{gao2013modified} put forward Evidential Semi-local Centrality (ESC) and pointed out that it is
more reasonable than the existing centrality measures
such as Degree Centrality (DC), Betweenness Centrality
(BC) and Closeness Centrality (CC). In the application of ESC, the degree and strength of each node are first expressed by bbas, and then the fused importance is calculated using the combination rule in belief function theory. The higher the
ESC value is, the more important the node is. The detail computation process of ESC can be found in  \citep{gao2013modified}.

The similarity or dissimilarity index signifies to
what extent the proximity between two  vertices of a graph is. The dissimilarity measure considered in this paper is the one put forward by \citet{zhou2003distance}.  This index relates a network to a discrete-time Markov chain and utilises the mean first-passage time to express the
distance between two nodes. One can refer to \citep{zhou2003distance} for more details.
\subsubsection{Modularity}
Recently, many criteria were proposed for evaluating the partition of a network. A widely used measure
called modularity, or $Q$ was presented by \citet{newman2004finding}.
Given a hard
partition with $c$ groups ($\omega_1,\omega_2,\cdots,\omega_c$) $\bm{U}=(u_{ik})_{n
\times c}$, where $u_{ik}$ is one if vertex $i$ belongs to the $k_{th}$
community, 0 otherwise, and let the $c$ crisp subsets of vertices be
$\{V_1,V_2,\cdots,V_c\}$,  then the modularity can be defined as \citep{fortunato2010community}:
 \begin{equation}\label{hard_modu2}
	Q_h=\frac{1}{\left \| \bm{W} \right \|}\sum_{k=1}^c \sum_{i,j \in V_k}
	\left(w_{ij}-\frac{k_i k_j}{\left \| \bm{W} \right \|}\right), \end{equation}
where $\left \| \bm{W} \right \|=\sum_{i,j=1}^n w_{ij}$, $k_i=\sum_{j=1}^n w_{ij}$.
The node subsets $\{V_k,k=1,2,\cdots,c\}$ are determined by the hard partition
$\bm{U}_{n \times c}$, but the role of $\bm{U}$ is somewhat obscured by this form of
modularity  function. To reveal the role played by the partition $\bm{U}$
explicitly, \citet{havens2013soft} rewrote the equations in the form of $\bm{U}$.
Let $\bm{k}=(k_1,k_2,\cdots,k_n)^{T}$, $\bm{B}=\bm{W}-\bm{k}^T\bm{k}/\left \| W \right
\|$, then \begin{align}\label{hard_modu3} Q_h&=\frac{1}{\left \| \bm{W} \right
	\|}\sum_{k=1}^c \sum_{i,j=1}^n \left(w_{ij}-\frac{k_i k_j}{\left \| \bm{W}
	\right \|}\right)u_{ik} u_{jk}\nonumber \\ &=\frac{1}{\left \| \bm{W} \right
	\|}\sum_{k=1}^c \bm{u}_k \bm{B} \bm{u}_k^\text{T}\nonumber \\ &=\text{trace}
	\left(\bm{U}^\text{T} \bm{B} \bm{U}\right)/ \|\bm{W}\|, \end{align} where
$\bm{u}_k=\left(u_{1k},u_{2k},\cdots,u_{nk}\right)^\text{T}$.

\citet{havens2013soft} pointed out that an advantage of
Eq.~\eqref{hard_modu3} is that it is well defined for any partition of the
nodes not just crisp ones. The fuzzy modularity of $\bm{U}$ was derived as
\begin{equation} Q_f=\text{trace} \left(\bm{U}^\text{T} \bm{B} \bm{U}\right)/ \|\bm{W}\|,
\end{equation} where $\bm{U}$ is the membership matrix and $u_{ik}$ represents the
membership of community $k$ for node $i$. If $u_{ik}$ is restricted in
$\left[0,1\right]$, the fuzzy partition degrades to the hard one, and so $Q_f$ equals to
$Q_h$ at this time.
\section{Median Evidential $C$-Means (MECM) approach}
We introduce here median evidential $c$-means in order to take advantages of both median clustering and
credal partitions. Like all the prototype-based clustering methods, for MECM,
an objective function should first be  found to provide an immediate measure of
the quality of clustering results. Our goal then can  be characterized as
the optimization of the objective function to get the best credal partition.
\subsection{The objective function of  MECM}
To group  $n$ objects in
$\bm{X}=\{\bm{x}_1,\bm{x}_2,\cdots,\bm{x}_n\}$ into $c$ clusters
$\omega_1,\omega_2,\cdots,\omega_c$, the credal partition
$\bm{M}=\{\bm{m}_1,\bm{m}_2,\cdots,\bm{m}_n\}$ defined on
$\Omega=\{\omega_1,\omega_2,\cdots,\omega_c\}$ is used to represent the class
membership of the objects, as in
\citep{denoeux2004evclus,masson2008ecm}.
The quantities $m_{ij}=m_i(A_j)$ $(A_j\neq \emptyset, A_j \subseteq \Omega)$
are determined by the dissimilarity between object $x_i$ and  focal set $A_j$
which has to be defined first.

Let the prototype set of specific (singleton) cluster be
$\bm{V}=\{\bm{v}_1,\bm{v}_2,\cdots,\bm{v}_c\}$, where $\bm{v}_i$ is the prototype
vector of  cluster $\omega_i$ $(i=1,2,\cdots,c)$ and it must be one of the $n$
objects. If $|A_j|=1$, {\em i.e.}, $A_j$ is associated with one of the
singleton clusters in $\Omega$ (suppose to be $\omega_j$ with prototype vector
$\bm{v}_j$), then the dissimilarity between $\bm{x}_i$ and $A_j$ is defined by
\begin{equation} \label{specific_pro}
	\overline{d}^2_{ij}=d^2(\bm{x}_i,\bm{v}_j), \end{equation} where
$d(\bm{x}_i,\bm{x}_j)$ represents the known dissimilarity between object
$\bm{x}_i$ and $\bm{x}_j$. When $|A_j|>1$, it represents an imprecise (meta)
cluster. If object $\bm{x}_i$ is to be partitioned into a meta cluster, two
conditions should be satisfied. One is  the dissimilarity values between
$\bm{x}_i$ and the included singleton classes'  prototypes are similar. The
other is the object should be close to   the prototypes of all these  specific
clusters.  The former measures the degree of  uncertainty, while the latter is
to avoid the pitfall of  partitioning two data objects irrelevant to any
included specific clusters into the corresponding imprecise classes.  Let the
prototype vector of the imprecise cluster  associated with $A_j$   be
$\overline{\bm{v}}_j$, then the dissimilarity between $\bm{x}_i$ and $A_j$ can
be defined as: \begin{align} \label{imprecise_pro}
	\overline{d}^2_{ij}&=\frac{\gamma \frac{1}{|A_j|} \sum\limits_{\omega_k
	\in A_j}d^2(\bm{x}_i,\bm{v}_k)+ \rho_{ij}
	\min\{d(\bm{x}_i,\bm{v}_k):\omega_k \in A_j\}}{\gamma +1}, \end{align}
with \begin{equation} \label{rhoij} \rho_{ij}=
	\frac{\sum\limits_{\omega_x,\omega_y \in
	A_j}\sqrt{(d\left(\bm{x}_i,\bm{v}_x)-d(\bm{x}_i,\bm{v}_y)\right)^2}}{ \eta
		\sum\limits_{\omega_x,\omega_y \in A_j}d(\bm{v}_x,\bm{v}_y) }.
	\end{equation} In Eq.~\eqref{imprecise_pro}
$\gamma$ weights the contribution of the dissimilarity of the objects from the
consisted specific clusters  and  it can be tuned according to the
applications. If $\gamma=0$, the imprecise clusters only consider our
uncertainty. Discounting factor $\rho_{ij}$ reflects the degree of
uncertainty. If $\rho_{ij}=0$, it means that all the dissimilarity values
between $\bm{x}_i$ and the included specific classes in $A_j$ are equal, and
we are absolutely uncertain about which cluster object $\bm{x}_i$ is actually
in. Parameter $\eta$ $(\in [0,1])$ can be tuned to control of the discounting
degree. 
In credal partitions, we
can distinguish between ``equal evidence" (uncertainty) and ``ignorance".  The
ignorance reflects the indistinguishability among the clusters. In fact,
imprecise classes take both uncertainty and ignorance into consideration, and
we can balance the two types of imprecise information by adjusting $\gamma$.
Therefore, the dissimilarity between $x_i$ and $A_j ( A_j \neq \emptyset, A_j
\subseteq \Omega)$, $\overline{d}_{ij}$, can be calculated by \begin{equation}
	\label{meta_pro} \overline{d}^2_{ij}=\begin{cases} d^2(\bm{x}_i,\bm{v}_j)
		& |A_j|=1, \\ \frac{\gamma \frac{1}{|A_j|} \sum\limits_{\omega_k \in
		A_j}d^2(\bm{x}_i,\bm{v}_k)+ \rho_{ij}
		\min\{d(\bm{x}_i,\bm{v}_k):\omega_k \in A_j\}}{\gamma +1} & |A_j|>1
	\end{cases}.  \end{equation}

Like ECM, we propose to look for the credal partition
$\bm{M}=\{\bm{m}_1,\bm{m}_2,\cdots,\bm{m}_n\}\in \mathcal{R}^{n\times 2^c}$ and the
prototype set $\bm{V}=\{\bm{v}_1,\bm{v}_2,\cdots,\bm{v}_c\}$ of specific (singleton) clusters by minimizing the objective
function: \begin{equation} J_{\mathrm{MECM}}(\bm{M},
	\bm{V})=\sum\limits_{i=1}^{n}\sum\limits_{A_j\subseteq \Omega,A_j \neq
	\emptyset}|A_j|^\alpha
	m_{ij}^{\beta}\overline{d}_{ij}^2+\sum\limits_{i=1}^{n}\delta^2m_{i\emptyset}^{\beta},
	\label{costfun} \end{equation} \noindent constrained on \begin{equation}
	\sum\limits_{A_j\subseteq \Omega,A_j \neq
	\emptyset}m_{ij}+m_{i\emptyset}=1, \label{MECMconstraint} \end{equation}
where $m_{ij}\triangleq m_{i}(A_j)$ is the bba of $n_i$ given to the nonempty
set $A_j$, $m_{i\emptyset} \triangleq m_{i}(\emptyset)$ is the bba of $n_i$
assigned to the empty set, and $\overline{d}_{ij}$ is the dissimilarity
between $x_i$ and  focal set $A_j$. Parameters $\alpha,\beta,\delta$ are adjustable with the same meanings
as those in ECM. Note that
$J_\text{MECM}$ depends on the credal partition $\bm{M}$ and the set $\bm{V}$ of all
prototypes.

\subsection{The optimization}
To minimize $J_\text{MECM}$, an optimization
scheme via an Expectation-Maximization (EM) algorithm  as in MCM
\citep{cottrell2006batch} and MFCM \citep{geweniger2010median} can be
designed, and the alternate update steps are as follows:

\noindent Step 1. Credal partition ($\bm{M}$) update.  \begin{itemize} \item
			$\forall A_j \subseteq \Omega, A_j \neq \emptyset$,
			\begin{equation} \label{mass1}
m_{ij}=\frac{|A_j|^{-\alpha/(\beta-1)}{\overline{d}_{ij}^{-2/(\beta-1)}}}
{\sum\limits_{A_k\neq\emptyset}|A_k|^{-\alpha/(\beta-1)}{\overline{d}_{ik}^{-2/(\beta-1)}}+\delta^{-2/(\beta-1)}}
		\end{equation}
\item if $A_j = \emptyset$, \begin{equation}
				\label{mass2} m_{i\emptyset}=1-\sum\limits_{A_j \neq
				\emptyset}m_{ij} \end{equation} \end{itemize} \noindent Step
2. Prototype ($\bm{V}$) update.

The prototype $v_i$ of a specific (singleton) cluster $\omega_i$
$(i=1,2,\cdots,c)$ can be updated first and then the dissimilarity between the
object and the prototype of each imprecise (meta) clusters associated with
subset $A_j \subseteq \Omega$ can be obtained by Eq.~\eqref{meta_pro}.  For
singleton clusters $\omega_k$ $(k=1,2,\cdots,c)$, the corresponding new
prototypes $v_k$ $(k=1,2,\cdots,c)$  are set to be sample $x_l$ orderly, with
\begin{equation}
\label{pro_update}
 x_l= \arg \min_{\bm{v}^{'}_k}
	\left\{L(\bm{v}^{'}_k) \triangleq \sum_{i=1}^n  \sum_{\omega_k \in A_j}
	|A_j|^\alpha  m_{ij}^\beta \overline{d}^2_{ij}(\bm{v}^{'}_k), \bm{v}^{'}_k
	\in \{\bm{x}_1,\bm{x}_2,\cdots,\bm{x}_n\}\right\},
\end{equation}
The dissimilarity between $\bm{x}_i$ and $A_j$, $\overline{d}^2_{ij}$,  is a function of $\bm{v}^{'}_k$, which is the
prototype of $\omega_k (\in A_j)$, and it should be one of the
$n$ objects in $\bm{X}=\{\bm{x}_1,\bm{x}_2,\cdots,\bm{x}_n\}$.

The bbas of the objects' class membership are updated  identically to ECM
\citep{masson2008ecm}, but it is worth noting that $\overline{d}_{ij}$ has
different meanings and less constraints as explained before. For the
prototype updating process the fact that the prototypes are assumed to be one
of the data objects is taken into account. Therefore, when the credal
partition matrix $\bm{M}$ is fixed, the new prototypes of the clusters can be
obtained in a simpler manner than in the case of ECM application. The MECM algorithm is summarized as Algorithm
\ref{alg:method}.
\begin{algorithm}\caption{\textbf{:} ~Median evidential
$c$-means  algorithm}\label{alg:method} \begin{algorithmic} \STATE
	{\textbf{Input} \hspace{1.8cm} dissimilarity matrix $\bm{D}\triangleq
	[d(x_i,x_j)]_{n\times n}$ for the $n$ objects
	$\{\bm{x}_1,\bm{x}_2,\cdots,\bm{x}_n\}$} \STATE {\textbf{Parameters}
	\hspace{0.9cm}$c$: number clusters $1<c<n$ \\ \hspace{3.01cm}$\alpha$:
	weighing exponent for cardinality \\ \hspace{3.01cm}$\beta >1$: weighting
	exponent \\ \hspace{3.01cm}$\delta>0$: dissimilarity between any object to
	the emptyset \\ \hspace{3.01cm}$\gamma>0$: weight of dissimilarity between
	data and  prototype vectors\\ \hspace{3.01cm}$\eta \in [0,1]$: control of the
	discounting degree\\
			   } \STATE {\textbf{Initialization} \hspace{0.5cm} Choose
			   randomly $c$ initial cluster prototypes from the objects
			   \hspace{3.01cm} \STATE {\textbf{Loop}}\hspace{2.0cm} $t
		   \leftarrow 0$} \\ \hspace{2.88cm} \textbf{Repeat} \STATE
\hspace{3.0cm}(1). $t\leftarrow t+1$\\ \hspace{3.0cm}(2). Compute $\bm{M}_t$ using
Eq.~\eqref{mass1}, Eq.~\eqref{mass2} and  $\bm{V}_{t-1}$\\ \hspace{3.0cm}(3).
Compute the new prototype set $\bm{V}_{t}$ using Eq.~\eqref{pro_update}\\
\hspace{2.88cm} \textbf{Until~} {the prototypes remain unchanged}
\end{algorithmic} \end{algorithm}

The convergence of MECM algorithm can be proved in the following lemma,
similar to the proof of median neural gas \citep{cottrell2006batch} and MFCM
\citep{geweniger2010median}.

\begin{lemma}
The MECM algorithm (Algorithm \ref{alg:method}) converges in a finite number of steps.
\begin{spacing}{1.5} \begin{proof}
            Suppose $\theta^{(t)}=(\bm{M}_t,\bm{V}_t)$ and $\theta^{(t+1)}=(\bm{M}_{t+1},\bm{V}_{t+1})$
			are the parameters from two successive iterations of MECM. We will
			first prove that
\begin{equation} \label{to_prove}
				J_{MECM}(\theta^{(t)})\geq J_{MECM}(\theta^{(t+1)}),
			\end{equation}
which shows MECM always monotonically decreases the objective function. Let
\begin{align} J^{(t)}_{MECM} &=
				\sum\limits_{i=1}^{n}\sum\limits_{A_j\subseteq \Omega,A_j \neq
				\emptyset}|A_j|^\alpha
				(m_{ij}^{(t)})^{\beta}(\overline{d}_{ij}^{(t)})^2+\sum\limits_{i=1}^{n}\delta^2
				(m_{i\emptyset}^{(t)})^{\beta} \nonumber \\ &\triangleq
				\sum_{i} \sum_{j} f_1(\bm{M}_t)f_2(\bm{V}_t)+\sum_{i}f_3(\bm{M}_t),
			\end{align}
            where $f_1(\bm{M}_t)=|A_j|^\alpha (m_{ij}^{(t)})^\beta$,
			$f_2(\bm{V}_t)=(d_{ij}^{(t)})^2$, and $f_3(\bm{M}_t)=\delta^2 (m_{i
			\emptyset}^{(t)})^\beta$.  $\bm{M}_{t+1}$ is then obtained by
			maximizing the right hand side of the equation above. Thus,
			\begin{align}
                 J^{(t)}_{MECM} & \geq   \sum_{i} \sum_{j}
				f_1(\bm{M}_{t+1})f_2(\bm{V}_t)+\sum_{i}f_3(\bm{M}_{t+1}) \label{ieq1}\\ &
				\geq  \sum_{i} \sum_{j}
				f_1(\bm{M}_{t+1})f_2(\bm{V}_{t+1})+\sum_{i}f_3(\bm{M}_{t+1}) \label{ieq2}\\ &
				=J^{(t+1)}_{MECM}.
            \end{align}
            This inequality ~\eqref{ieq1}
			comes from the fact $\bm{M}_{t+1}$ is determined by differentiating of
			the respective Lagrangian of the cost function with respect to
			$\bm{M}_{t}$. To get Eq.~\eqref{ieq2}, we could use the fact that every
			prototype $\bm{v}_k$ $(k=1,2,\cdots,c)$ in $\bm{V}_{t+1}$ is orderly
			chosen explicitly to be $$\arg \min_{\bm{v}^{'}_k}
			\left\{L(v^{'}_k) \triangleq \sum_{i=1}^n  \sum_{\omega_k \in A_j}
			|A_j|^\alpha  m_{ij}^\beta \overline{d}^2_{ij}(\bm{v}^{'}_k),
			\bm{v}^{'}_k \in \{\bm{x}_1,\bm{x}_2,\cdots,\bm{x}_n\}\right\}, $$
			and thus this formula evaluated at $\bm{V}_{t+1}$ must be equal to or
			less than the same formula evaluated at $V_{t}$.

Hence MECM causes the objective function to converge monotonically. Moreover,
the bba $\bm{M}$ is a function of the prototypes $\bm{V}$ and for given $\bm{V}$ the
assignment $\bm{M}$ is unique. Because MECM assumes that the prototypes are
original object data in  $\bm{X}$, so there is a finite number of
different prototype vectors $\bm{V}$ and so is the number of corresponding credal
partitions $\bm{M}$. Consequently we can get the conclusion that the MECM algorithm
converges in a finite number of steps.
\end{proof}\end{spacing} \end{lemma}

\textbf{Remark 1.} Although the objective function of MECM takes the same form as that in ECM
\citep{masson2008ecm}, we should note that in MECM, it is no
longer assumed that there is an underlying Euclidean distance. Thus the dissimilarity measure $d_{ij}$ has few
restrictions such as the  triangle inequality or the symmetry. This freedom distinguishes the MECM from ECM and RECM, and it
leads to the constraint for the prototypes to be data
objects themselves. The distinct  difference in the process of minimization between MECM and ECM lies in the prototype-update step.
The purpose of updating the prototypes is to make sure that the cost function would decrease. In ECM the Lagrange multiplier optimization is
evoked directly while in MECM a search method is applied. As a result, the objective
function may decline more quickly in ECM as the optimization process has few constraints. However, when the centers of clusters in the data set
are more likely to be
the data object, MECM may converge with few steps.

\textbf{Remark 2.} Although  both MECM and MFCM can be applied to the same type of data set,  they are very different. This is due to
the fact that they are founded on
different models of partitioning. MFCM provides fuzzy partition. In contrast, MECM gives credal partitions. We emphasize that MECM is in line
with MCM and MFCM: each class is represented by a prototype which is restricted to the data objects
and the dissimilarities are not assumed to be fulfilling any metric properties. MECM is an extension of MCM and MFCM in the framework of belief functions.
\subsection{The parameters of the algorithm} As in ECM, before running MECM,
the values of the parameters have to be set. Parameters $\alpha, \beta$ and
$\delta$ have the same meanings as those in ECM, and $\gamma$ weighs the
contribution of  uncertainty to the dissimilarity between nodes and imprecise
clusters. The value $\beta$ can be set to be $\beta=2$ in all experiments for which it
is a usual choice. The parameter $\alpha$  aims to penalize the subsets with
high cardinality and control  the amount of points assigned to imprecise
clusters in both ECM and  MECM. As the measures for the dissimilarity between
nodes and meta classes are different, thus different values of $\alpha$ should
be taken even for the same data set. But both in ECM and MECM, the higher
$\alpha$ is, the less mass belief is assigned to the meta clusters and the
less imprecise will be the resulting partition. However, the decrease of
imprecision may result in high risk of errors. For instance, in the case of hard
partitions, the clustering results are completely precise but there is much more intendancy  to partition an object to
an unrelated group. As suggested in ~\citep{masson2008ecm}, a value can be used as a
starting default one but it can be modified according to what is expected from
the user.  The choice $\delta$ is more difficult and is strongly data
dependent~\citep{masson2008ecm}.

For determining the number of clusters, the validity index  of a credal
partition  defined by  \citet{masson2008ecm}
could be utilised:
\begin{equation}\label{clu_num_ind} N^*(c)\triangleq \frac{1}{n \log_2(c)}
	\times \sum_{i=1}^n \left[ \sum_{A\in 2^\Omega \setminus \emptyset}
	m_i(A)\log_2|A|+m_i(\emptyset)\log_2(c)\right], \end{equation}
where $0 \leq N^*(c) \leq 1$. This index has to be minimized to get the optimal number
of clusters. When MECM is applied to community detection, a different index is
defined to determine the number of communities. We will describe it in the
next section.
\section{Application and evaluation issues}
In this section, we
will discuss how to apply MECM to community detection problems in social
networks and how to evaluate credal partitions.
\subsection{Evidential modular function}
Assume the obtained credal partition of the graph is
$$\bm{M}=\left[\bm{m}_{1},\bm{m}_{2},\cdots,\bm{m}_{n}\right]^\text{T},$$ where
$\bm{m}_i=(m_{i1},m_{i2},\cdots,m_{ic})^{\text{T}}$. Similarly to the fuzzy
modularity  by  Havens et al.~\cite{havens2013soft}, here we introduce an
evidential modularity \citep{zhou2014evidential}: \begin{equation}\label{evi_modu} Q_e=\frac{1}{\left \|
	\bm{W} \right \|}\sum_{k=1}^c \sum_{i,j=1}^n \left(w_{ij}-\frac{k_i k_j}{\left
	\| \bm{W} \right \|}\right)pl_{ik} pl_{jk}, \end{equation} where
$\bm{pl}_{i}=\left(pl_{i1},pl_{i2},\cdots,pl_{ic}\right)^\text{T}$ is the
contour function associated with $m_i$, which describes the upper value of our
belief to the proposition that the $i_{th}$ node belongs to the $k_{th}$
community.

Let $\bm{PL}=(pl_{ik})_{n\times c}$, then Eq.~\eqref{evi_modu} can be rewritten as:

\begin{equation}\label{evi_modu1}
	Q_e=\frac{\mathrm{trace}(\bm{PL}^\text{T}~\bm{B}~\bm{PL})}{\left \| \bm{W} \right \|}.
\end{equation} $Q_e$ is a directly extension of the crisp  and fuzzy
modularity functions in Eq.~\eqref{hard_modu3}. When the credal partition
degrades into the hard and fuzzy ones, $Q_e$ is equal to $Q_h$ and $Q_f$
respectively.

\subsection{The initial prototypes for communities} \label{seed_choose}
Generally speaking, the person who is the center in the community in a social
network has the following characteristics: he has relation with all the
members of the group and their relationship is stronger than usual; he  may
directly contact with other persons who also play an important role in their
own community. For instance, in Twitter
network, all the members in the community of the fans of Real Madrid football
Club (RMC) are following the official account of the team, and RMC must be the
center of this community. RMC follows the famous football player in the club,
who is sure to be the center of the community of his fans. In fact,
RMC has 10382777 followers and 30 followings (the data on March 16, 2014).
Most of the followings have more than 500000 followers. Therefore, the centers
of the community can be set to the ones not only with high degree and strength,
but also with neighbours  who also have high degree and strength. Thanks to the theory
of belief functions, the evidential semi-local centrality  ranks the
nodes considering all these measures. Therefore the initial $c$ prototypes of
each community can be set to the nodes with largest ESC values.

Note that there is usually more than one center in one community. Take Twitter
network for example again, the fans of RMC who follow the club official account may
also pay attention to Cristiano Ronaldo, the most popular player in the team,
who could be  another center of the community of RMC's fans
to a great extent.
These two centers (the accounts of the club and Ronaldo) both have large ESC values but they are near to each other.
This situation violates the rule
which requires the chosen seeds as far away from each other as possible
\citep{arthur2007k,jiang2012efficient}.

The dissimilarity between the nodes could be utilised to solve this problem.
Suppose the ranking order of the nodes with respect to their ESCs is $n_1\geq
n_2 \geq \cdots \geq n_n$. In the beginning $n_1$ is set to be the first
prototype as it has the largest ESC, and then node $n_2$ is considered. If
$d(n_1,n_2)$ (the dissimilarity between node 1 and 2) is larger than a
threshold $\mu$, it is chosen to be the second prototype. Otherwise, we abandon
$n_2$ and turn to check $n_3$.  The process continues until all the $c$
prototypes are found. If there are not enough prototypes after checking all the nodes,
we should decrease $\mu$ moderately and restart the search from
$n_1$. In this paper we test the approach with the dissimilarity measure proposed in \citep{zhou2003distance}. 
Based on our experiments, $[0.7,1]$ is a better experiential range of the threshold $\mu$. This seed choosing strategy is similar to that in
\citep{jiang2012efficient}.

\subsection{The community detection algorithm based on MECM} The whole
community detection  algorithm in social networks based on MECM  is summarized
in Algorithm \ref{alg:method_MECM_community}.
\begin{algorithm}\caption{\textbf{:} ~Community detection algorithm based on
MECM}\label{alg:method_MECM_community} \begin{algorithmic}
 \STATE{\textbf{Input:}  $\bm{A}$, the adjacency matrix; $\bm{W}$, the weight matrix (if
	any); $\mu$, the threshold controlling the dissimilarity between the
	prototypes; $c_\text{min}$, the minimal number of communities;
	$c_{\text{max}}$, the maximal of communities; the required
parameters in original MECM algorithm}
\STATE {\textbf{Initialization:}
Calculate the dissimilarity matrix of the nodes in the graph.} \REPEAT
\STATE (1). Set the cluster number $c$ in MECM
be $c=c_\text{min}$.\\
(2). Choose the initial $c$ prototypes using the
strategy proposed in section~\ref{seed_choose}.\\
 (3). Run MECM with the
corresponding parameters and the initial prototypes got in (2).\\
(4). Calculate the evidential modularity using Eq.~\eqref{evi_modu1}.\\
 (5). Let $c=c+1$.  \UNTIL{$c$ reaches at $c_\text{max}$.}
 \STATE {\textbf{Output:}
Choose the number of communities at around which the modular function peaks,
and  output the corresponding credal partition of the graph.}
\end{algorithmic} \end{algorithm}

In the algorithm, $c_\text{min}$ and $c_\text{max}$ can be determined based on the
original graph. Note that $c_\text{min}\geq 2$. It is an empirical range of the
community number of the network. If $c$ is given, we can get a credal
partition based on MECM and then the evidential modularity  can be derived. As
we can see, the modularity is a function of $c$ and it should peak at around
the optimal value of $c$ for the given network.

\subsection{Performance evaluation} \label{evulation}
The objective of the
clustering problem is to partition a similar data pair to the same group.
There are two types of correct decisions by the clustering result: a true
positive (TP) decision assigns two similar objects to the same cluster, while
a true negative (TN) decision assigns two dissimilar objects to different
clusters. Correspondingly, there are two types of errors we can commit: a
false positive (FP) decision assigns two dissimilar objects to the same
cluster, while a false negative (FN) decision assigns two similar objects to
different clusters. Let $a$ (respectively, $b$) be the number of pairs of
objects simultaneously assigned to identical classes (respectively, different
classes) by the stand reference partition and the obtained one. Actually $a$
(respectively, $b$) is the number of TP (respectively, TN)  decisions.
Similarly, let $c$ and $d$ be the numbers of FP and FN decisions respectively.
Two  popular  measures   that  are  typically  used  to  evaluate the
performance  of hard clusterings  are  precision  and  recall. Precision (P)  is the fraction  of relevant
instances (pairs in identical groups in  the clustering benchmark)  out  of
those  retrieved  instances (pairs in identical groups of the discovered
clusters), while recall (R)  is the fraction of relevant instances that are
retrieved. Then precision and recall can be calculated by
 \begin{equation}
	\label{precision} \text{P}=\frac{a}{a+c}
	~~~~~\text{and}~~~~~\text{R}=\frac{a}{a+d}
 \end{equation} respectively.
The Rand index (RI) measures the percentage of correct decisions and it can be defined as
\begin{equation} \label{ri}
	\text{RI}=
	\frac{2(a+b)}{n(n-1)},
 \end{equation}
where $n$ is the number of data
objects. 
In fact, precision measures the
rate of the first type of errors (FP),  recall (R) measures another type (FN),
while RI measures both.

For fuzzy and evidential clusterings,  objects  may  be partitioned into
multiple  clusters with different degrees.  In  such cases precision  would
be  consequently  low \citep{mendes2003evaluating}.  Usually the  fuzzy and
evidential clusters  are  made  crisp  before calculating  the  measures,
using  for  instance  the  maximum membership criterion
\citep{mendes2003evaluating} and pignistic probabilities
\citep{masson2008ecm}. Thus in the work  presented in this paper, we have
hardened the fuzzy and credal clusters by maximizing the corresponding
membership and pignistic probabilities and calculate precision, recall and RI
for each case.

The introduced imprecise clusters can avoid the risk to group a data into a
specific class without strong belief. In other words, a data pair can be
clustered into the same specific group only when we are quite confident and
thus the misclassification rate will be reduced. However, partitioning too
many data into  imprecise clusters may cause that  many objects are not
identified for their precise groups. In order to show the effectiveness of the
proposed method in these aspects, we use the evidential precision (EP) and
evidential recall (ER):
\begin{equation}\label{ep}
	\text{EP}=\frac{n_{er}}{N_e}, ~~~ \text{ER}=\frac{n_{er}}{N_r}.
\end{equation}
In Eq.~\eqref{ep}, the notation $N_e$ denotes the number of
pairs partitioned into the same specific group by evidential clusterings, and
$n_{er}$ is the number of  relevant instance pairs out of these specifically
clustered pairs. The value $N_r$  denotes the number of pairs in the same
group of the clustering benchmark, and ER is the fraction of  specifically
retrieved instances (grouped into an identical specific cluster) out of these
relevant pairs. When the partition degrades to a crisp one, EP and ER equal to
the classical precision and recall measures respectively.  EP and ER reflect
the accuracy of the credal partition from different points of view, but we
could not evaluate the clusterings from one single term. For example, if all
the objects are partitioned into imprecise clusters except two relevant data
object grouped into a specific class, $\text{EP}=1$ in this case. But we could
not say this is a good partition since it does not provide us with any information
of great value. In this case $\text{ER}\approx 0$. Thus ER could be used to express
the efficiency of the method for providing valuable partitions. Certainly we can combine EP and ER
like RI to get the evidential rank index (ERI) describing the accuracy:
\begin{equation}
\label{eri} \text{ERI}=\frac{2(a^*+b^*)}{n(n-1)},
\end{equation} where  $a^*$ (respectively, $b^*$) is the number of pairs of
objects simultaneously clustered to the same specific class ({\em i.e.},
singleton class, respectively, different classes) by the stand reference
partition and the obtained credal one.
Note that for  evidential  clusterings, precision, recall and RI  measures are
calculated after the corresponding hard partitions are got, while EP, ER and
ERI are based on hard credal partitions \citep{masson2008ecm}.

\begin{center} \begin{figure}[!thbt] \centering
 \includegraphics[width=0.45\linewidth]{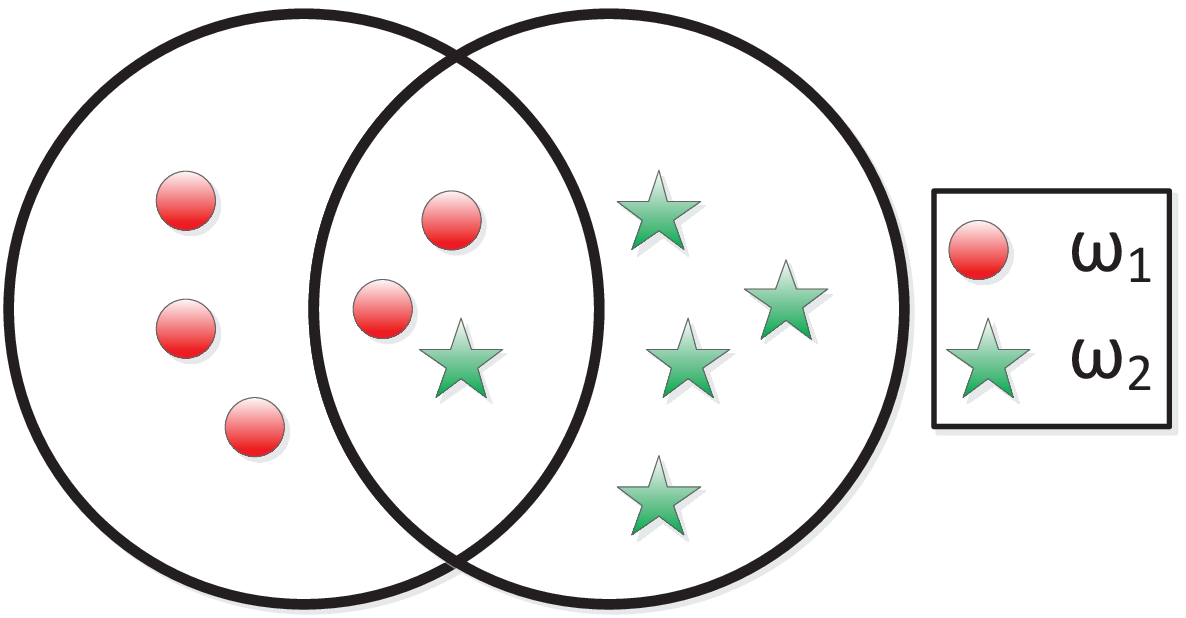}
\hfill \includegraphics[width=0.45\linewidth]{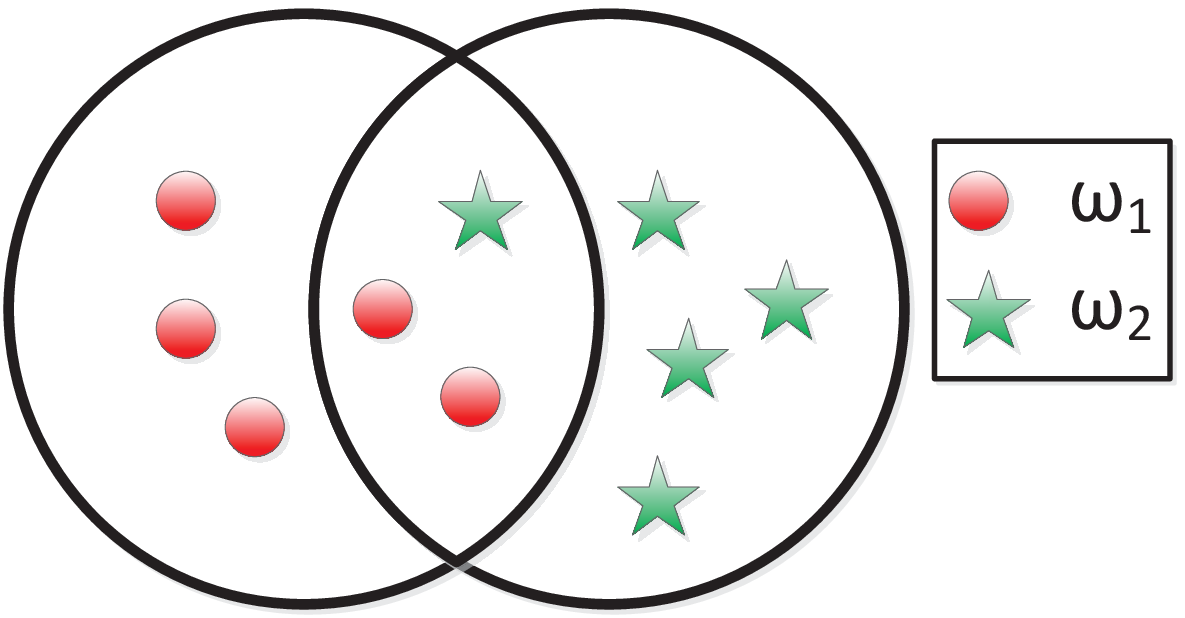}
		\hfill
		\parbox{.45\linewidth}{\centering\small a. Original data sets.}\hfill
\hfill \parbox{.45\linewidth}{\centering\small b. Partition 1.}		
\includegraphics[width=0.45\linewidth]{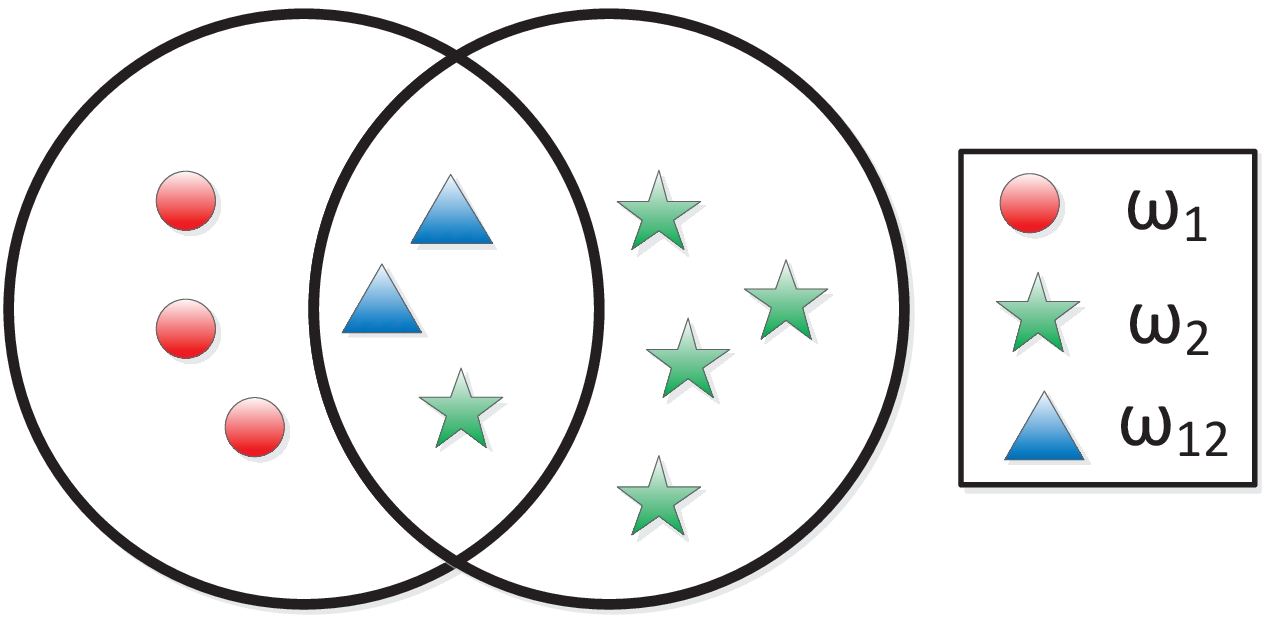}
		\hfill
		\includegraphics[width=0.45\linewidth]{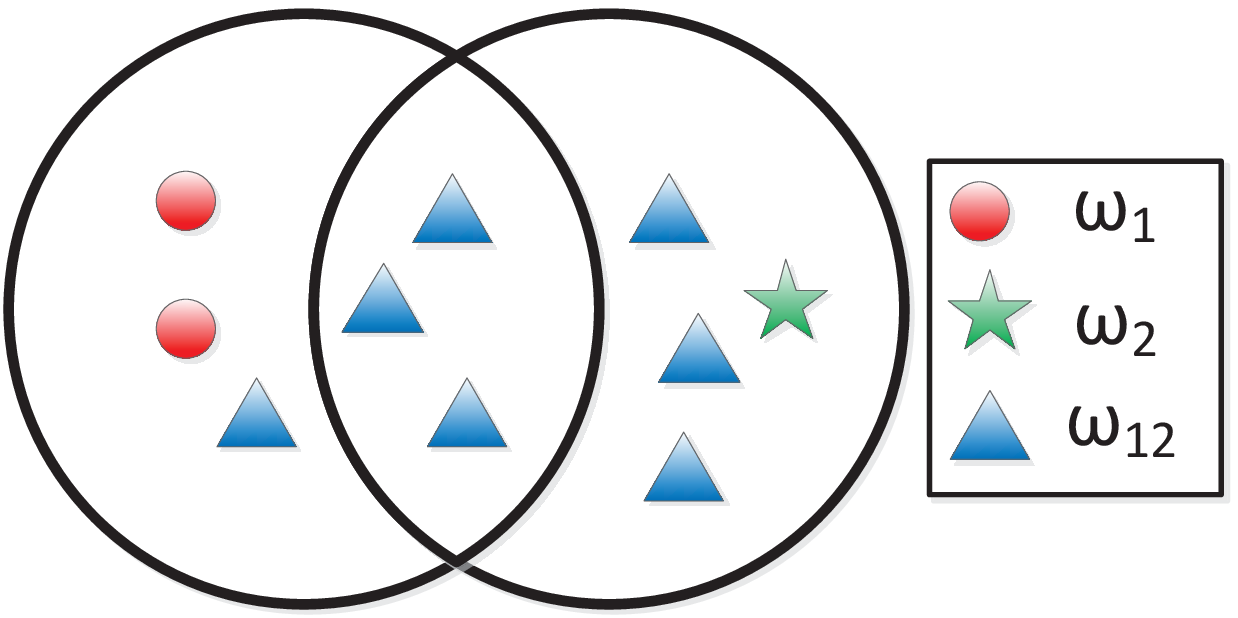}
		\hfill \parbox{.45\linewidth}{\centering\small c. Partition 2.}
		\hfill \parbox{.45\linewidth}{\centering\small d. Partition 3.}
\hfill \caption{An small data set with imprecise classes.} \label{cluster} \end{figure} \end{center}

\textbf{Example 1.} In order to show the significance of the above performance measures, an example containing only ten objects from two groups is presented here. The three partitions are given in Fig.~\ref{cluster}-b -- \ref{cluster}-d. The values of the six evidential indices (P,R,RI,EP,ER,ERI) are listed in Tab.~\ref{cluster_table}.
\begin{table}[ht]
\centering
\caption{Evaluation Indices of the partitions.}
\begin{tabular}{rrrrrrr}
  \hline
 & EP & ER & ERI & P & R & RI \\
  \hline
  Partition 1& 0.6190 & 0.6190 & 0.6444& 0.6190 & 0.6190 & 0.6444 \\
  Partition 2 & 1.0000 & 0.6190 & 0.8222&--&--&-- \\
  Partition 3 & 1.0000 & 0.0476 & 0.5556&--&--&-- \\
   \hline
\end{tabular}
\label{cluster_table}
\end{table}
We can see that if we simply partition the nodes in the overlapped area, the risk of misclassification is high in terms of precision. The introduced imprecise cluster $\omega_{12}  \triangleq \{\omega_1,\omega_2\}$ could enable us to make soft decisions, as a result the accuracy of the specific partitions is high. However, if too many objects are clustered into imprecise classes, which is  the case of partition 3, it is pointless although EP is high. Generally, EP denotes the accuracy of the specific decisions, while ER represents the efficiency of the approach. We remark that the evidential indices degrade to the corresponding classical indices ({\em e.g,} evidential precision degrades to precision) when the partition is crisp.
\section{Experiments} \label{sec_tests}
In this section a number of experiments are performed on classical data sets in the distance space and on
graph data for which  only the dissimilarities between nodes are known. The
obtained credal partitions are compared with hard and fuzzy ones using the
evaluation indices proposed in Section~\ref{evulation} to show the merits of
MECM.
\subsection{Overlapped data set} Clustering approaches to detect overlap objects
which leads to recent attentions are still inefficiently processed. Due to the introduction of imprecise
classes, MECM has the advantage to detect overlapped clusters. In the first example, we will
use  overlapped data sets to illustrate the behavior of the proposed algorithm.

\begin{center}
	\begin{figure}[!thbt] \centering
		\includegraphics[width=0.45\linewidth]{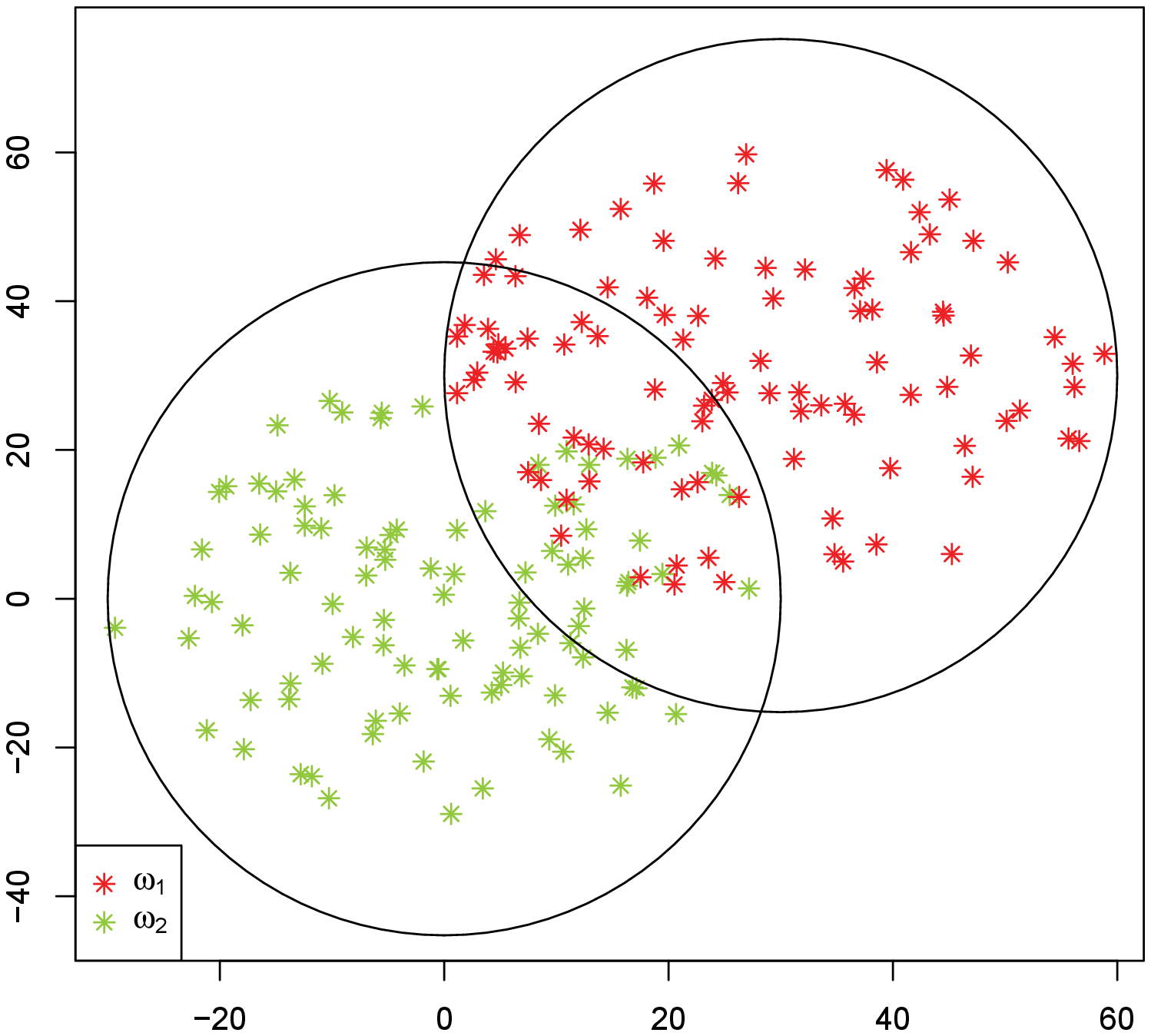}
		\hfill
		\includegraphics[width=0.45\linewidth]{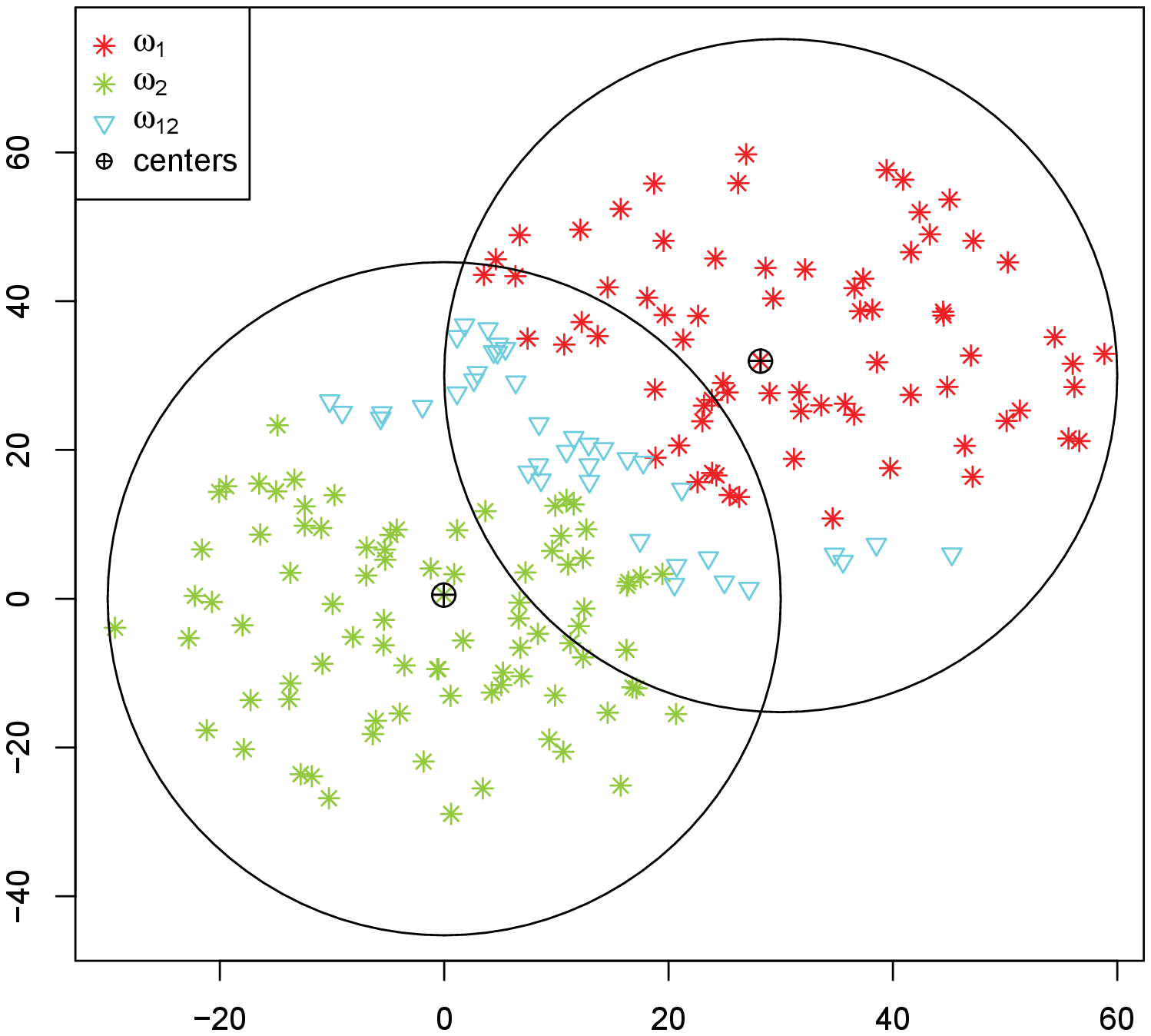}
		\hfill \parbox{.45\linewidth}{\centering\small a. Overlapped data}
		\hfill \parbox{.45\linewidth}{\centering\small b. Clustering result of
		MECM} \hfill \caption{Clustering of overlapped data set}
		\label{rcircle_data} \end{figure} \end{center}

We start by generating  $2 \times 100$ points uniformly distributed in two overlapped
circles with a same radius $R=30$ but with different  centers. The coordinates of
the first circle's center are $(0,0)$ while the coordinates of the other circle's center are $(30,30)$. The data
set is displayed in Fig.~\ref{rcircle_data}-a.  

In order to show the influence of parameters in MECM and ECM, different values
of $\gamma$, $\alpha$, $\eta$ and $\delta$ have been tested for this data set.
The figure Fig.~\ref{diffpara}-a displays the three evidential indices varying with
$\gamma$ ($\alpha$ is fixed to be 2) by MECM, while Fig.~\ref{diffpara}-b
depicts the results of MECM with different $\alpha$ but a fixed  $\gamma=0.4$
($\eta$ and $\delta$ are set 0.7 and 50, respectively,  in the tests). For
fixed $\alpha$ and $\gamma$, the results with different $\eta$ and $\delta$
are shown in Fig.~\ref{diffpara}-c. The effect of $\alpha$ and $\delta$ on the
clusterings of ECM  is illustrated in Fig.~\ref{diffpara}-e. As we can see,
for both MECM and ECM, if we want to make more imprecise decisions to improve
ER, parameter $\alpha$ can be decreased. In MECM, we can also reduce the value of parameter $\gamma$ to
accomplish the same purpose.
Although both $\alpha$ and $\gamma$
have effect on imprecise clusters in MECM, the mechanisms they work are
different. Parameter $\alpha$ tries to adjust the penalty degree to control
the imprecise rates of the results. However, for $\gamma$, the same aim could be
got by regulating the uncertainty degree of imprecise classes. It can be
seen from the figures, the effect of $\gamma$ is more conspicuous than
$\alpha$. Moreover, although $\alpha$ may be set too high to obtain good
clusterings, ``good"  partitions can also be got by adjusting $\gamma$ in this
case. For both MECM and ECM, the stable limiting values of evidential
measures are around 0.7 and 0.8. Such values suggest the equivalence of the two
methods to a certain  extent.  Parameter $\eta$ is used for discounting the
distance between uncertain objects and specific clusters. As pointed out in
Fig.~\ref{diffpara}-c, if $\gamma$ and $\alpha$ are well set, it has little
effect on the final clusterings. The same is true in the case of $\delta$ which is applied to detect outliers. The effect of the different values of parameter $\beta$ is
illustrated in Fig.~\ref{diffpara}-d. We can see that it has little influence on the
final results as long as it is larger than 1.  As in FCM and ECM, for which it is a usual
choice, we use $\beta=2$ in all the following experiments.

The improvement  of precision will bring about the
decline of recall, as more data could not be clustered into specific classes.
What we should do is to set parameters based on our own requirement to make a
tradeoff between precision and recall. For instance, if we want to make a
cautious decision in which EP is relatively high, we can reduce $\gamma$ and
$\alpha$. 
Values of these parameters can be also learned from  historical data if such data are available.
\begin{center} \begin{figure}[!thbt] \centering
		\hfill
		\hfill \parbox{.45\linewidth}{\centering\small a. MECM (with respect
		to $\gamma$)} \hfill \parbox{.45\linewidth}{\centering\small b. MECM
		(with respect to $\alpha$)} \hfill
		\hfill
		\hfill \parbox{.45\linewidth}{\centering\small c. MECM (with respect
		to $\eta$ or $\delta$)} \hfill \parbox{.45\linewidth}{\centering\small
		d. MECM (with respect to $\beta$)} \hfill
        \parbox{.45\linewidth}{\centering\small
		e. ECM (with respect to $\alpha$ or $\delta$)}
        \caption{Clustering of overlapped data set with different parameters}
		\label{diffpara} \end{figure} \end{center}

For the objects in the overlapped area, it is difficult to make a hard decision {\em i.e.} to decide about their
specific groups. Thanks to the imprecise clusters, we can make a soft decision.  As analysed before, the soft decision
will improve the precision of total results and reduce the risk of
misclassifications caused by simply partitioning  the overlapped objects into specific class. However, too many imprecise decisions will decrease the recall value. Therefore, the  ideal partition should make a compromise between the two measures.  Set $\alpha=1.8,\gamma=0.2,\eta=0.7$ and $\delta=50$, the ``best" (with relatively high values on both precision and recall)  clustering result by MECM is shown in Fig.~\ref{rcircle_data}-b. As we can see, most of the data in
the overlapped area are partitioned into imprecise cluster
$\omega_{12}\triangleq\{\omega_1,\omega_2\}$ by the application of MECM. We adjust the coordinates  of the
center of the second circle to get overlapped data with different proportions
(overlap rates), and the validity indices of the clustering results  by
different methods are illustrated in Fig.~\ref{barplot}. For the application of MECM, MCM and MFCM, each algorithm is evoked 20 times with randomly selected initial prototypes for the same data set and the mean values of the evaluating indices are reported. The figure Fig.~\ref{barplot}-d shows the average values of the indices by MECM (plus and minus one standard deviation) for 20 repeated experiments as a function of the overlap rates. As we can see the initial prototypes indeed have effects on the final results, especially when the overlap rates are high. Certainly, we can avoid the influence by repeating the algorithm many times. But this is too expensive for MECM. Therefore, we suggest to use the prototypes obtained in MFCM or MCM as the initial. In the following experiments, we will set the initial prototypes to be the ones got by MFCM.

As it can be seen, for different overlap rates, the classical measures such as
precision, recall, and RI are almost the same for all the methods. This
reflects that  pignistic probabilities  play a similar role as fuzzy
membership. But we can see that for MECM,  EP is significantly high, and the
increasing of overlap rates has least effects on it compared with the other
methods. Such effect can be attributed to the introduced imprecise clusters which
enable us to make a compromise decision between hard ones. But as many
points are clustered into imprecise classes, the evidential recall value is
low.

Overall, this example reflects one of the superiority of MECM that it can detect overlapped clusters.
The objects in the overlapped area could be clustered into imprecise classes by this approach.  Other possible available
information or special
techniques could be utilised for these imprecise data when we have to make hard decisions.  Moreover,
partitions with different degree of  imprecision can be got by
adjusting the parameters of the algorithm based on our own requirement.

 \begin{center} \begin{figure}[!thbt] \centering
		 \hfill
		 \hfill \parbox{.45\linewidth}{\centering\small a. Precision} \hfill
		 \parbox{.45\linewidth}{\centering\small b. Recall} \hfill
		 \hfill
          \parbox{.45\linewidth}{\centering\small c. RI} \hfill
          \parbox{.45\linewidth}{\centering\small d. Repeated experiments for MECM} \hfill
		 \caption{Clustering of overlapped data set with different overlap
		 rates. Figure-d shows the average values of the indices (plus and minus one standard deviation) for 20 repeated experiments, as a function of the overlap rates. For MECM $\alpha=1.8,\gamma=0.2,\eta=0.7,\delta=50$.}
		 \label{barplot} \end{figure} \end{center}

\subsection{Classical data sets from Gaussian mixture model}
In the second experiment, we test on a data set consisting of $3 \times 50+2 \times 5$ points generated from different Gaussian distributions. The first
$3\times 50$ points are from Gaussian distributions $G(\bm{\mu}_k,\bm{\Sigma}_k)
(k=1,2,3)$ with \begin{align} \bm{\mu}_1=&(0, 0)^\text{T},
	\bm{\mu}_2=(40,40)^\text{T},\bm{\mu}_3=(80,80)^\text{T}\\ \nonumber
	&\bm{\Sigma}_1=\bm{\Sigma}_2=\bm{\Sigma}_3=\begin{pmatrix} 120 & 0 \\ 0 & 120
	\end{pmatrix}, \end{align} and the last $2 \times 10 $ data are noisy
points follow  $G(\bm{\mu}_k,\bm{\Sigma}_k) (k=4,5)$ with
\begin{align} \bm{\mu}_4=&(-50, 90)^\text{T}, \bm{\mu}_5=(-10,130)^\text{T}\\
	\nonumber & \bm{\Sigma}_4=\bm{\Sigma}_5=\begin{pmatrix} 80 & 0 \\ 0 & 80
	\end{pmatrix}.  \end{align}
MECM is applied with the following settings:
$\alpha=1, \delta=100, \eta=0.7$, while ECM has been tested using
$\alpha=1.7,\delta=100$ (The appropriate parameters can be determined similarly as in the first
example).  One can see from Figs.~\ref{gaussian_data}-b and
\ref{gaussian_data}-c , MCM and MFCM can partition most of the regular data in
$\omega_1$, $\omega_2$ and $\omega_3$ into their correct clusters, but they
could not detect the noisy points correctly. These noisy data are simply grouped into a specific
cluster by both approaches. As can be seen from Fig.~\ref{gaussian_data}-d, for the points
located in the middle part of $\omega_2$,  ECM could not find their exact group and misclassify
them into imprecise cluster $\omega_{13}$ . In the figures
$\omega_{ij}\triangleq\{\omega_i,\omega_j\}$ denotes imprecise clusters.

As mentioned before, imprecise classes in MECM can measure ignorance and
uncertainty at the same time, and the degree of ignorance in meta clusters can
be adjusted by $\gamma$. We can see that MECM does not detect
many points in the overlapped area between two groups if $\gamma$ is set to 0.6.
In such a case the test objects are
partitioned into imprecise clusters mainly because of our ignorance about their
specific classes. These objects attributed to meta classes mainly belong to noisy data in
$\omega_4$ and $\omega_5$. The distance of these points to the prototypes of
specific clusters is large (but not too large or they could be regarded to be
in the emptyset, see Fig.~\ref{gaussian_data}-e). Thus the distance between
the prototype vectors is relatively small so that these specific clusters are
indistinguishable.  Decreasing $\gamma$ to be 0.2 would make imprecise class denoting
more uncertainty, as it can be seen from Fig.~\ref{gaussian_data}-f, where many
points located in the margin of each group are clustered into imprecise
classes. In such a case, meta classes rather reflect our uncertainty on the data
objects' specific cluster.

The table Tab.~\ref{precision_gaussian_point} lists the indices for evaluating the
different methods. Bold entries in each column of this table (and also other tables in the following) indicate
that the results are significant as the top performing
algorithm(s) in terms of the corresponding evaluation index. We can see that the precision, recall and RI
values for all approaches are similar except from those obtained for ECM which are significantly
lower.  As these classical measures are based on the associated pignistic
probabilities for evidential clusterings, it seems that credal partitions can
provide the same information as crisp and fuzzy ones. But from the same table, we
can also see that the evidential measures EP and ERI obtained for MECM are higher (for hard
partitions, the values of evidential measures equal  to the  corresponding
classical ones) than  the ones obtained for other methods. This fact confirms the accuracy of the specific decisions
{\em i.e.} decisions clustering the objects into specific classes. The advantage can be attributed to
the introduction of imprecise clusters, with which we do not have to partition the
uncertain or unknown objects into a specific cluster. Consequently, it could reduce the risk of
misclassification. However, although ECM also deals with imprecise
clusters, the accuracy is not improved as much as in the case of applying MECM. As illustrated
before in the case of ECM application, many objects of a specific cluster are partitioned into an
irrelevant imprecise class and, as a result, the evidential precision value
and ERI decrease as well.

\begin{center} \begin{figure}[!thbt] \centering
		\hfill
		\hfill \parbox{.45\linewidth}{\centering\small a. Original artificial
		data} \hfill \parbox{.45\linewidth}{\centering\small b. Clustering
		result of MCM} \hfill
		\hfill
		\hfill \parbox{.45\linewidth}{\centering\small c. Clustering result of
		MFCM} \hfill \parbox{.45\linewidth}{\centering\small d. Clustering
		result of ECM} \hfill
		\hfill
		\hfill \parbox{.45\linewidth}{\centering\small e. Clustering result of
		MECM ($\gamma=0.6$)} \hfill \parbox{.45\linewidth}{\centering\small f.
		Clustering result of MECM ($\gamma=0.2$)} \hfill \caption{Clustering
		of the  artificial data set by different methods}
		\label{gaussian_data} \end{figure} \end{center}
\begin{table}[ht] \centering \caption{The  clustering results for gaussian
	points  by different methods. For each method, we generate 20 data sets with the same parameters and report the mean values of the
  evaluation indices for all the data sets.
} \begin{tabular}{rrrrrrrr} \hline & Precision  & Recall & RI & EP & ER &ERI\\
	\hline
MCM   & 0.7802 & 0.9570 & 0.9002  & 0.7802 & 0.9570 & 0.9002\\
MFCM  & 0.8616 & 0.9797 & 0.9484 & 0.8616 & 0.9797 & 0.9484\\
FCM   & 0.8644 & 0.9820 & 0.9500 & 0.8644 & \textbf{0.9820} & 0.9500\\
ECM   & 0.8215 & 0.9353 & 0.9222 & 0.9069 & 0.8436 & 0.9294 \\
MECM ($\gamma=0.2$)  & \textbf{0.8674} & \textbf{0.9855} & \textbf{0.9520} & \textbf{0.9993} & 0.7721 & 0.9336  \\
MECM ($\gamma=0.6$)  & 0.8662 & 0.9851 & 0.9515 & 0.9958 & 0.9586 & \textbf{0.9868} \\
\hline \end{tabular}
\label{precision_gaussian_point} \end{table}

We also test on ``Iris flower", ``cat cortex"  and ``protein" data sets
\citep{fisher1936use,graepel1999classification,hofmann1997pairwise}. The first is object data while
the other two are relational data sets. Thus we compare our method with FCM and ECM  for the Iris data set, and with
RECM and NRFCM (Non-Euclidean Relational Fuzzy Clustering Method \citep{hathaway1994nerf}) for the last two data sets. The
results are displayed in Fig.~\ref{different_datasets}.

Presented results allow us to sum up the characteristics of MECM. Firstly, one can see that the behavior of MECM is similar to ECM
for traditional data. Besides, credal partitions provided by MECM allow to recover
the information of crisp and fuzzy partitions. Moreover, we are able to balance influence of our
uncertainty and ignorance according to the actual needs. The
examples utilised before deal with classical data sets. But the superiority of MECM
makes it applicable in the case of data sets for which only dissimilarity
measures are known {\em e.g.} social networks. Thus in the following experiments, we will use
some graph data to illustrate the behaviour of the proposed method on the
community detection problem in social networks. The dissimilarity index used here is the one brought forward by \citet{zhou2003distance}. To have a
fair comparison, in the following experiments, we also compare with three classical algorithms for community detection  {\em i.e.} 
BGLL \citep{blondel2008fast}, 
LPA \citep{raghavan2007near} and  ZFCM (a fuzzy $c$-means based approach proposed by \citet{zhang2007identification}).  The obtained community structures are compared with known performance measures, {\em i.e.,} NMI (Normalized Mutual Information), VI (Variation of Information) and Modularity.
\begin{center}
	\begin{figure}[!thbt] \centering
		\hfill
		\hfill \parbox{.45\linewidth}{\centering\small a. Precision} \hfill
		\parbox{.45\linewidth}{\centering\small b. Recall} \hfill
		\hfill \parbox{.8\linewidth}{\centering\small c. RI} \hfill
		\caption{Clustering results of different data sets}
		\label{different_datasets} \end{figure} \end{center}

\subsection{Artificial graphs and generated benchmarks}
 To show the performance of the algorithm in detecting communities in networks, we first
apply the method to a sample network generated from Gaussian mixture model.
This model has been used for testing community detection approaches by \citet{liu2010detecting}.

The artificial graph is composed of $3\times 50$ nodes,
$n_1,n_2,\cdots,n_{150}$, which are represented  by 150 sample points,
$\bm{x}_1, \bm{x}_2, \cdots, \bm{x}_{150}$, in two-dimensional Euclidean
space. There are $3 \times 50$ points generated from Gaussian distributions
$G(\bm{\mu}_k, \bm{\Sigma}_k) (k=1,2,3)$ with \begin{align} \bm{\mu}_1=(1.0,&
	4.0)^\text{T},
	\bm{\mu}_2=(2.5,5.5)^\text{T},\bm{\mu}_3=(0.5,6.0)^\text{T}\\ \nonumber
	&\bm{\Sigma}_1=\bm{\Sigma}_2=\bm{\Sigma}_3=\begin{pmatrix} 0.25 & 0 \\ 0 & 0.25
	\end{pmatrix}.  \end{align}
Then, the edges of the graph are generated by
the following thresholding strategy: if $|x_i-x_j|\leq dist $, we set an edge
between node $i$ and node $j$; Otherwise the two nodes are not directly
connected. The graph is shown in Fig.~\ref{gaussian_network}-a (with
$dist=0.8$) and the dissimilarity matrix of the nodes is displayed in
Fig.~\ref{gaussian_network}-b. From the figures we can see that there are three
significant communities in the graph, and some nodes in the  bordering of
their groups seem to be in overlapped classes as they contact with members in
different communities simultaneously.
The table Tab.~\ref{precision_gaussian} lists the indices for evaluating the results. It shows that MECM performs well as the evidential
precision resulting from its application is high. MECM utilization  also results
in decreasing the probabilities of clustering failure thanks to the introduction of imprecise clusters.
This makes the decision-making process more cautious and reasonable.
\begin{center} \begin{figure}[!thbt] \centering
		\hfill
		\hfill \parbox{.45\linewidth}{\centering\small a. Gaussian  network}
		\hfill \parbox{.45\linewidth}{\centering\small b. Dissimilarity
		matrix} \hfill \caption{Artificial network from Gaussian mixture
		model} \label{gaussian_network} \end{figure} \end{center}
\begin{table}[ht]\small \centering \caption{The results for Gaussian graph by
	different methods} \begin{tabular}{rrrrrrrrrrr} \hline
 & Precision  & Recall & RI & EP & ER &ERI & NMI & VI & Modularity\\ \hline
 MCM      & 0.9049 & 0.9110 & 0.9392  & 0.9049 & 0.9110 & 0.9392& 0.8282 & 0.3769 & 0.6100 \\
 MFCM   & 0.9067 & 0.9099 & 0.9396  &0.9067 & 0.9099 & 0.9396  & 0.8172 & 0.4013 & 0.6115 \\
 ZFCM    & 0.9202 & 0.9224 & 0.9482 &0.9202 & 0.9224 & 0.9482 & 0.8386 & 0.3545 & 0.6118 \\
 MECM    & \textbf{0.9470} & \textbf{0.9472} & \textbf{0.9652} & \textbf{0.9789} & 0.6060 & 0.8661 & \textbf{0.8895} & \textbf{0.2428} & 0.6072 \\
  BGLL & 0.9329 & 0.9347 & 0.9564 & 0.9329 & 0.9347 & \textbf{0.9564} & 0.8597 & 0.3081 & \textbf{0.6119} \\
LPA     & 0.3289 & 1.0000 & 0.3289 & 0.3289 & 1.0000 & 0.3289 & 0.0000 & 1.0986 & 0.0000 \\
 \hline \end{tabular} \label{precision_gaussian}
 \end{table}

The algorithms are also compared by means of \citet{lancichinetti2008benchmark} benchmark (LFR) networks. The results of different methods in two kinds of LFR networks with 500 and 1000 nodes are displayed in Figs.~\ref{LFR500}--\ref{LFR1000} respectively.  The parameter $\mu$ showed in the $x$-axis in the figures identifies  whether the network has clear communities. When $\mu$ is small, the graph has well community structure. In such a case, almost all the methods perform well. But we can see that when $\mu$ is large, the results by MECM have the  largest values of precision. It means that the decisions which partition the nodes into a specific cluster are of great confidence. In terms of NMI, the results are similar to those by BGLL and LPA, but better than those of MCM and MFCM. This fact well explains that the hard or fuzzy partitions could be recovered when necessary.
 \begin{center} \begin{figure}[!thbt] \centering
		 \hfill
		 \hfill \parbox{.45\linewidth}{\centering\small a. Precision} \hfill
		 \parbox{.45\linewidth}{\centering\small b. NMI} \hfill
		 \caption{Comparison of MECM and other algorithms in LFR networks. The number of nodes is $N=500$.  The average degree is $|k| = 15$, and the pair for the exponents is $(\gamma, \beta)=(2,1)$.}
		 \label{LFR500} \end{figure} \end{center}

\begin{center} \begin{figure}[!thbt] \centering
		 \hfill
		 \hfill \parbox{.45\linewidth}{\centering\small a. Precision} \hfill
		 \parbox{.45\linewidth}{\centering\small b. NMI} \hfill
		 \caption{Comparison of MECM and other algorithms in LFR networks. The number of nodes is $N=1000$.  The average degree is $|k| = 20$, and the pair for the exponents is $(\gamma, \beta)=(2,1)$.}
		 \label{LFR1000} \end{figure} \end{center}
\subsection{Some real-world networks}
 \textbf{A. Zachary's karate club.}
 The Zachary's Karate Club data \citep{zachary1977information} is an
 undirected graph which consists of 34 vertices and 78 edges. 
 The original  graph and the dissimilarity of  the nodes are shown in
 Fig.~\ref{karate_ori}-a and \ref{karate_ori}-b respectively.

Let the parameters of MECM be $\alpha=1.5, \delta=100, \eta=0.9,\gamma=0.6$.
The modularity functions by MECM, MCM, MFCM and ZFCM
(Fig.~\ref{modularity_fun_karate}-a) peak around $c=2$ and $c=3$. Let $c=2$,
all the methods  can detect the two communities exactly.  If we set $c=3$, a
small community, which can also be found in the dissimilarity matrix
(Fig.~\ref{karate_ori}-b),  is separated from $\omega_1$  by
all the approaches (see Fig.~\ref{karate_ori_result}). But ZFCM assigns the maximum membership to
$\omega_1$ for node 9, which is actually in $\omega_2$. It seems  that
the loss of accuracy in the mapping process may cause such results.

MECM does not find imprecise groups when $\gamma=0.6$ as the network has
apparent community structure, and this reflects the fact that the communities
are distinguishable for all the nodes. But there may be some overlap between
two communities. The nodes in the overlapped cluster can be detected by
decreasing $\gamma$ (increasing the  uncertainty for imprecise communities).
As is  displayed in Fig.~\ref{karate_ori_result}-c and d, by declining
$\gamma$ to 0.1 and 0.05 respectively (the other parameters remain unchanged),
nodes 3 and 9 are clustered into both $\omega_1$ and $\omega_2$ $(\omega_{12})$
one after another.

From the results we can see that MECM takes both the ignorance and the
uncertainty into consideration while introducing imprecise communities. The degree of ignorance and
uncertainty could be balanced through adjusting $\gamma$. The analysis shows that there appears only uncertainty without ignorance in the original
club network. In order to show
the performance of MECM when there are noisy conditions such that some
communities are indistinguishable, two noisy nodes are added to the original
graph in the next experiment.
  \begin{center} \begin{figure}[!thbt] \centering
		\hfill
		\hfill \parbox{.45\linewidth}{\centering\small a. Original karate club
		network} \hfill \parbox{.45\linewidth}{\centering\small b.
		Dissimilarity matrix} \hfill \caption{Original karate club network}
		\label{karate_ori} \end{figure} \end{center} \begin{center}
	\begin{figure}[!thbt] \centering
		\hfill
		\hfill \parbox{.45\linewidth}{\centering\small a. Clustering result of
		ZFCM} \hfill \parbox{.45\linewidth}{\centering\small b. Clustering
		result of MCM, MFCM, and MECM ($\gamma=0.1$)} \hfill
		\hfill
		\hfill \parbox{.45\linewidth}{\centering\small c. Clustering result of
		MECM ($\gamma=0.1$)} \hfill \parbox{.45\linewidth}{\centering\small d.
		Clustering result of MECM ($\gamma=0.05$)} \hfill \caption{Detected
		communities of karate club network by different methods}
		\label{karate_ori_result} \end{figure} \end{center}
   \begin{center}
	\begin{figure}[!thbt] \centering
		\hfill
		\hfill \parbox{.45\linewidth}{\centering\small a. Original karate
		club} \hfill \parbox{.45\linewidth}{\centering\small b. Karate club
		with noisy nodes} \hfill \caption{Modularity functions of karate club
		network by different methods} \label{modularity_fun_karate}
	\end{figure} \end{center}

\textbf{B. Karate club network with some added noisy nodes.}
In this test, two noisy nodes are added to the original
karate club network (see Fig.\ref{karate_add}-a). The first one is node 35,
which is directly connected with nodes 18 and 27. The other one is 36, which
is connected to nodes 1 and 33. It can be seen from the dissimilarity
matrix that node 36 has stronger relationships  with both communities than
node 35. This is due to the fact that the nodes connected to node 36
play leader roles in their own group, but node 35 contacts with two
marginal nodes with ``small" or insignificant  roles in their own group only.

The results obtained by the application of different methods are shown in
Fig.~\ref{karate_add_result}. The MECM parameters  are set as follows:
$\alpha=1.5, \delta=100, \eta=0.9$ and $\gamma$ is tuned according to the
the extent that the imprecise communities reflect our
ignorance. As we can see, MCM, MFCM and ZFCM simply group the two noisy nodes
into $\omega_1$. With $\gamma=0.4$ MECM regards node 36 as a member of $\omega_1$
while node 35 is grouped into  imprecise community $\omega_{12}$. And
$\omega_{12}$ mainly reflects our ignorance rather than uncertainty on the actual
community of node 36. This is why node 36 is not clustered into
$\omega_{12}$ since $\omega_1$ and $\omega_2$ are distinguishable for him but
we are just not sure for the final decision. The increase in the extent of
uncertainty in imprecise communities results from the decrease of  $\gamma$ value. We can see that
more nodes (including nodes 36,9,1,12,27, see Figs.~\ref{karate_add_result}-e
and f) are clustered into $\omega_{12}$ or $\omega_{13}$ due to uncertainty.
The imprecise communities consider both ignorance (node 35) and
uncertainty (other nodes).

These results reflect the difference between ignorance and uncertainty. As
node 35 is only related to one outward node of each  community, thus we are
ignorant about which community it really belongs to.  On the contrary,
node 36 connects with the key members (playing an important role in the
community), and in this case the dissimilarity between the prototypes of
$\omega_1$ and $\omega_2$ is relatively large so they are distinguishable.
Thus there is uncertainty rather than ignorance about which community node 36
is in. In this network, node 36 is a ``good" member for both communities,
whereas node 35 is a ``poor" member.  It can be seen from
Fig.~\ref{karate_add_mass_member}-a that the  fuzzy partition  by MFCM also gives
large similar membership values to $\omega_1$ and $\omega_2$ for node 35, just
like in the case of such good members as node 36 and 9. The obtained results show  the problem  of
distinguishing between ignorance and the ``equal evidence" (uncertainty) for fuzzy
partitions. But Fig.~\ref{karate_add_mass_member}-b shows that the credal
partition by MECM assigns small mass belief to $\omega_1$ and $\omega_2$ for
node 35, indicating  our ignorance on its situation.
\begin{center} \begin{figure}[!thbt] \centering
		\hfill
		\hfill \parbox{.45\linewidth}{\centering\small a.  Karate club network
		with added nodes} \hfill \parbox{.45\linewidth}{\centering\small b.
		Dissimilarity matrix} \hfill \caption{Karate club network with two
		noisy nodes} \label{karate_add} \end{figure} \end{center}
\begin{center} \begin{figure}[!thbt] \centering
		\hfill
		\hfill \parbox{.45\linewidth}{\centering\small a.  MCM} \hfill
		\parbox{.45\linewidth}{\centering\small b. ZFCM} \hfill
		\hfill
		\hfill \parbox{.45\linewidth}{\centering\small c.  MFCM} \hfill
		\parbox{.45\linewidth}{\centering\small d. MECM $(\gamma=0.4)$} \hfill
		\hfill
		\hfill \parbox{.45\linewidth}{\centering\small e.  MECM
		$(\gamma=0.1)$} \hfill \parbox{.45\linewidth}{\centering\small f. MECM
		$(\gamma=0.02)$} \hfill \caption{ Detected communities in Karate club
		network with noisy nodes} \label{karate_add_result} \end{figure}
\end{center}

 \begin{center} \begin{figure}[!thbt] \centering
		\hfill
		\hfill \parbox{.45\linewidth}{\centering\small a. Fuzzy membership by
		MFCM} \hfill \parbox{.45\linewidth}{\centering\small b. Mass belief by
		MECM} \hfill \caption{Fuzzy membership and mass belief of the nodes in
		karate club network with noisy nodes } \label{karate_add_mass_member}
	\end{figure} \end{center}
We also test our method on four other real-world graphs: American football network, Dolphins network, Lesmis network and Political
books network\footnote{These data sets can be found in http://networkdata.ics.uci.edu/index.php}. 
The measures applied to evaluate the performance of different methods are listed in Tab.~\ref{precision_football}--\ref{polbooks}. It can been seen from the tables, for all the graphs MECM application results in a community structure with high evidential precision level. The precision results from a cautious decision making process which clusters the noisy nodes into imprecise communities. In terms of classical performance measures like NMI, VI and modularity, MECM slightly outperforms the other algorithms. Note that  these classical measures for hard partitions are calculated by the pignistic probabilities associated with the credal partitions provided by MECM. Therefore, we can also see the possibility to recover the hard decisions here when using the proposed evidential detection approach.

\begin{table}[ht] \small \centering \caption{The
	results for American football network by different methods}
	\begin{tabular}{rrrrrrrrrrr} \hline
  & Precision  & Recall & RI & EP & ER &ERI &NMI &VI & Modularity\\ \hline
 MCM  & 0.7416 & 0.8834 & 0.9661 & 0.7416 & 0.8834 & 0.9661 & 0.8637 & 0.6467 & 0.5862\\
 MFCM  & 0.7583 & 0.8757 & 0.9678 & 0.7583 & 0.8757 & 0.9678 & 0.8715 & 0.6160 & 0.5745\\
 ZFCM  & 0.8176 & 0.9082 & 0.9765 & 0.8176 & 0.9082 & 0.9765  & 0.9035 & 0.4653 & 0.6022\\
 MECM & \textbf{0.8232} & 0.9082 & \textbf{0.9771} & \textbf{0.9303} & 0.8681 & \textbf{0.9843} & \textbf{0.9042} & \textbf{0.4625} & 0.5995\\
 BGLL   & 0.7512 & \textbf{0.9120} & 0.9689 & 0.7512 & \textbf{0.9120} & 0.9689 & 0.8903 & 0.5195 & \textbf{0.6046} \\
 LPA    & 0.6698 & 0.8298 & 0.9538 & 0.6698 & 0.8298 & 0.9538 & 0.8623 & 0.6580 & 0.5757 \\
  \hline \end{tabular} \label{precision_football}
  \end{table}

\begin{table}[ht] \small \centering \caption{The
	results for Dolphins network by different methods}
	\begin{tabular}{rrrrrrrrrrr} \hline
  & Precision  & Recall & RI & EP & ER &ERI &NMI &VI & Modularity\\ \hline
 MCM   & 1& 1 & 1 & 1 & 1 & 1 & 1 & 0 & 0.3787\\
 MFCM  & 1& 1 & 1 & 1 & 1 & 1 & 1 & 0 & 0.3787\\
 ZFCM   & 1& 1 & 1 & 1 & 1 & 1 & 1 & 0 & 0.3787\\
 MECM  & 1& 1 & 1 & 1 & 1 & 1 & 1 & 0 & 0.3787\\
   BGLL   & 0.9271 & 0.3583 & 0.6351& 0.9271 & 0.3583 & 0.6351  & 0.4617 & 1.1784 & 0.5185 \\
LPA   & 0.9250 & 0.5029 & 0.7070 & 0.9250 & 0.5029 & 0.7070 & 0.5595 & 0.8354 & 0.5070    \\
  \hline \end{tabular} \label{dolphins}
  \end{table}

\begin{table}[ht]\small \centering \caption{The
	results for Lesmis network by different methods}
	\begin{tabular}{rrrrrrrrrrr} \hline
  & Precision  & Recall & RI & EP & ER &ERI &NMI &VI & Modularity\\ \hline
 MCM   & 0.6109 & 0.5522 & 0.9005 & 0.6109 & 0.5522 & 0.9005 & 0.7381 & 1.1295 & 0.4732 \\
 MFCM  & 0.5774 & 0.6456 & 0.8971 & 0.5774 & 0.6456 & 0.8971 & 0.7743 & 0.9555 & 0.4705\\
 ZFCM  & \textbf{0.7368} & 0.5769 & 0.9217  & 0.7368 & 0.5769 & 0.9217 & 0.7805 & 0.9666 & 0.4983 \\
 MECM  &  0.7065 & 0.7473 & \textbf{0.9299} & \textbf{0.9298} & 0.4368 & \textbf{0.9258} & \textbf{0.7977} & \textbf{0.8531} & 0.4884 \\
   BGLL   & 0.5796 & 0.8104 & 0.9033  & 0.5796 & 0.8104 & 0.9033 & 0.7551 & 0.9435 & \textbf{0.5556}\\
 LPA  & 0.4594 & \textbf{0.9643} & 0.8544 & 0.4594 & \textbf{0.9643} & 0.8544 & 0.7500 & 0.8637 & 0.5428    \\
\hline \end{tabular} \label{lesmis}
  \end{table}

\begin{table}[ht]\small \centering \caption{The
	results for Political books network by different methods}
	\begin{tabular}{rrrrrrrrrrr} \hline
  & Precision  & Recall & RI & EP & ER &ERI &NMI &VI & Modularity\\ \hline
 MCM   & 0.8109 & 0.8030 & 0.8482 & 0.8109 & 0.8030 & 0.8482  & 0.5721 & 0.8426 & 0.4979 \\
 MFCM  & 0.8020 & 0.8187 & \textbf{0.8485} & 0.8020 & 0.8187 & \textbf{0.8485}  & \textbf{0.5755} & 0.8256 & 0.4962 \\
 ZFCM  & 0.7928 & 0.7487 & 0.8234  & 0.7928 & 0.7487 & 0.8234 & 0.5301 & 0.9407 & 0.5048 \\
 MECM  & 0.7880 & 0.8081 & 0.8383 & \textbf{0.8458} & 0.6435 & 0.8128 & \textbf{0.5755} & 0.8247 & 0.4725 \\
   BGLL   & \textbf{0.8244} & 0.6203 & 0.7978 & 0.8244 & 0.6203 & 0.7978  & 0.5121 & 1.0987 & \textbf{0.5205}\\
 LPA  & 0.7331 & \textbf{0.8558} & 0.8200 & 0.7331 & \textbf{0.8558} & 0.8200  & 0.5612 & \textbf{0.7925} & 0.4604   \\
\hline \end{tabular} \label{polbooks}
  \end{table}
\subsection{Discussion}
We will discuss for which application MECM is designed here. As analysed before, for MECM only dissimilarities between objects  are required and only the intuitive assumptions need to be satisfied for the dissimilarity measure.  Therefore, the
algorithm could be  appropriate for many clustering tasks for non-metric
data objects. This type of data is  very common in social sciences, psychology, etc, where any metric assumptions about the
similarities/dissimilarities could not be assured. The freedom for the data set leads to the restriction that the prototypes should be the
objects themselves.  Nevertheless, this constraint seems reasonable for social networks as the center of a community is usually the
person (node) frequently contacting with others.  Thus the approach can be applied to community detection problems. Thanks to the introduction of imprecise classes, it could reduce the risk of partitioning the objects which we are uncertain or ignorant   into an incorrect cluster.  For this reason the algorithm can help us make soft decisions when clustering  the data set without distinct cluster/community structures or with overlap.

Due to the computational complexity, the proposed  algorithm is  not well directly adapted to handle very large data sets. However,  here we discuss the possibility to apply the evidential community detection approach to large-scale networks.  Firstly, the number of parameters to be optimized is exponential and depends on the number of clusters \citep{masson2008ecm}. For the number of classes larger than
10, calculations are not tractable. But we can consider only a subclass with a
limited number of focal sets \citep{masson2008ecm}. For instance, we could
constrain the focal sets to be composed of at most two classes (except
$\Omega$). Secondly, for the network with millions of nodes, MCM or MFCM could be evoked as a first step
to merge some nodes into small clusters. After that we can apply MECM to the ``coarsened" network. But how to define the edges or connections of the new graph should be studied. Lastly we  emphasize that the evidential community detection algorithm could be utilised for gaining a better insight into the network structure and detecting the imprecise classes. For the large-scale network, it is difficult to make specific decisions for all of nodes due to the limitation of time,  money or techniques. In this case we can use the proposed approach to make some ``soft" decisions first and then  use some  techniques  special for  the imprecise parts of the graph.
\section{Conclusion}
  We introduced a Median variant of Evidential $C$-means (MECM) as a new prototype-based
  clustering algorithm in the present contribution. The proposed approach is an extension of median $c$-means and
  median fuzzy $c$-means. It is based on the framework of belief function theory. The
  applied median-based  clustering requires the definition of the dissimilarity between the objects only. Therefore, it is not
  restricted to a metric space application. The prototypes of the clusters are
  constrained  to the data objects themselves. MECM provides us with not only credal partitions  but also hard and fuzzy partitions
  as by-products through computing pignistic probabilities. Moreover, it could distinguish ignorance
  from uncertainty while the fuzzy or crisp partitions could not. By the
  introduced imprecise clusters, we could find some overlapped and
  indistinguishable clusters for related nodes. Thanks to the advantages of belief function theory
  and median clustering, MECM could be  applied to community detection
  problems in social networks. As other median clustering approaches, MECM
  tends to get stuck in local minima such that several runs have to be
  performed to obtain good performance. However, we propose an initial
  prototype-selection scheme using the evidential semi-centrality for
  the application of MECM in community detection to solve the
  problems brought by the initial prototypes. Results of presented experiments on artificial and
  real-world networks show that the  credal partitions on graphs provided by
  MECM application are more refined than crisp and fuzzy ones. Therefore, they could enable us to
  gain a better understanding of analysed community structure. Some examples on the
  classical metric space are also given to illustrate the interest of MECM and
  to show its difference with respect to the existing methods.

  As mentioned in this paper, there may be more than one center in each community
  network. Nevertheless, we ignore ``multi-center" to avoid the troubles
  brought by the need for an initial seed using ESC and the definition of a threshold to control the
  distance between prototypes. We are aware that this is a drawback of the presented approach  as not all
  the centers in each community are taken into consideration. Therefore, we intend to include the feature of
  multiple centers in our future research work.

\section*{Acknowledgements}
The authors are grateful to the anonymous reviewers for all
their remarks which helped us to clarify and improve the quality
of this paper. This work was supported by the National
Natural Science Foundation of China (Nos.61135001, 61403310).
\bibliographystyle{elsarticle-harv}
\addcontentsline{toc}{section}{\refname}\bibliography{paperlist}

\begin{thebibliography}{50}
\expandafter\ifx\csname natexlab\endcsname\relax\def\natexlab#1{#1}\fi
\expandafter\ifx\csname url\endcsname\relax
  \def\url#1{\texttt{#1}}\fi
\expandafter\ifx\csname urlprefix\endcsname\relax\def\urlprefix{URL }\fi

\bibitem[{Amiri et~al.(2013)Amiri, Hossain, Crawford, and
  Wigand}]{amiri2013community}
Amiri, B., Hossain, L., Crawford, J.~W., Wigand, R.~T., 2013. Community
  detection in complex networks: Multi--objective enhanced firefly algorithm.
  Knowledge-Based Systems 46, 1--11.

\bibitem[{Arthur and Vassilvitskii(2007)}]{arthur2007k}
Arthur, D., Vassilvitskii, S., 2007. $k$-means++: The advantages of careful
  seeding. In: Proceedings of the eighteenth annual ACM-SIAM symposium on
  Discrete algorithms. Society for Industrial and Applied Mathematics, pp.
  1027--1035.

\bibitem[{Bezdek(1981)}]{bezdek1981pattern}
Bezdek, J.~C., 1981. Pattern recognition with fuzzy objective function
  algorithms. Kluwer Academic Publishers.

\bibitem[{Blondel et~al.(2008)Blondel, Guillaume, Lambiotte, and
  Lefebvre}]{blondel2008fast}
Blondel, V.~D., Guillaume, J.-L., Lambiotte, R., Lefebvre, E., 2008. Fast
  unfolding of communities in large networks. Journal of Statistical Mechanics:
  Theory and Experiment 2008~(10), P10008.

\bibitem[{Bordogna and Pasi(2012)}]{bordogna2012quality}
Bordogna, G., Pasi, G., 2012. A quality driven hierarchical data divisive soft
  clustering for information retrieval. Knowledge-based systems 26, 9--19.

\bibitem[{Borgelt(2006)}]{borgelt2006prototype}
Borgelt, C., 2006. Prototype-based classification and clustering. Ph.D. thesis,
  Otto-von-Guericke-Universit{\"a}t Magdeburg, Universit{\"a}tsbibliothek.

\bibitem[{Clauset et~al.(2004)Clauset, Newman, and Moore}]{clauset2004finding}
Clauset, A., Newman, M.~E., Moore, C., 2004. Finding community structure in
  very large networks. Physical review E 70~(6), 066111.

\bibitem[{Cottrell et~al.(2006)Cottrell, Hammer, Hasenfu{\ss}, and
  Villmann}]{cottrell2006batch}
Cottrell, M., Hammer, B., Hasenfu{\ss}, A., Villmann, T., 2006. Batch and
  median neural gas. Neural Networks 19~(6), 762--771.

\bibitem[{Den{\oe}ux and Masson(2004)}]{denoeux2004evclus}
Den{\oe}ux, T., Masson, M.-H., 2004. {EVCLUS}: evidential clustering of
  proximity data. Systems, Man, and Cybernetics, Part B: Cybernetics, IEEE
  Transactions on 34~(1), 95--109.

\bibitem[{Duch and Arenas(2005)}]{duch2005community}
Duch, J., Arenas, A., 2005. Community detection in complex networks using
  extremal optimization. Physical review E 72~(2), 027104.

\bibitem[{Dunn(1973)}]{dunn1973fuzzy}
Dunn, J.~C., 1973. A fuzzy relative of the isodata process and its use in
  detecting compact well-separated clusters. Journal of Cybernetics 3, 32--57.

\bibitem[{Fisher(1936)}]{fisher1936use}
Fisher, R.~A., 1936. The use of multiple measurements in taxonomic problems.
  Annals of eugenics 7~(2), 179--188.

\bibitem[{Fortunato(2010)}]{fortunato2010community}
Fortunato, S., 2010. Community detection in graphs. Physics Reports 486~(3),
  75--174.

\bibitem[{Fortunato and Barthelemy(2007)}]{fortunato2007resolution}
Fortunato, S., Barthelemy, M., 2007. Resolution limit in community detection.
  Proceedings of the National Academy of Sciences 104~(1), 36--41.

\bibitem[{Gao et~al.(2013)Gao, Wei, Hu, Mahadevan, and Deng}]{gao2013modified}
Gao, C., Wei, D., Hu, Y., Mahadevan, S., Deng, Y., 2013. A modified evidential
  methodology of identifying influential nodes in weighted networks. Physica A:
  Statistical Mechanics and its Applications 392~(21), 5490--5500.

\bibitem[{Geweniger et~al.(2010)Geweniger, Z{\"u}lke, Hammer, and
  Villmann}]{geweniger2010median}
Geweniger, T., Z{\"u}lke, D., Hammer, B., Villmann, T., 2010. Median fuzzy
  c-means for clustering dissimilarity data. Neurocomputing 73~(7), 1109--1116.

\bibitem[{Graepel et~al.(1999)Graepel, Herbrich, Bollmann-Sdorra, and
  Obermayer}]{graepel1999classification}
Graepel, T., Herbrich, R., Bollmann-Sdorra, P., Obermayer, K., 1999.
  Classification on pairwise proximity data. Advances in neural information
  processing systems, 438--444.

\bibitem[{Hammer and Hasenfuss(2007)}]{hammer2007relational}
Hammer, B., Hasenfuss, A., 2007. Relational neural gas. In: KI 2007: Advances
  in Artificial Intelligence. Springer, pp. 190--204.

\bibitem[{Hathaway and Bezdek(1994)}]{hathaway1994nerf}
Hathaway, R.~J., Bezdek, J.~C., 1994. Nerf $c$-means: Non-euclidean relational
  fuzzy clustering. Pattern recognition 27~(3), 429--437.

\bibitem[{Hathaway et~al.(1989)Hathaway, Davenport, and
  Bezdek}]{hathaway1989relational}
Hathaway, R.~J., Davenport, J.~W., Bezdek, J.~C., 1989. Relational duals of the
  $c$-means clustering algorithms. Pattern recognition 22~(2), 205--212.

\bibitem[{Havens et~al.(2013)Havens, Bezdek, Leckie, Ramamohanarao, and
  Palaniswami}]{havens2013soft}
Havens, T., Bezdek, J., Leckie, C., Ramamohanarao, K., Palaniswami, M., 2013. A
  soft modularity function for detecting fuzzy communities in social networks.
  Fuzzy Systems, IEEE Transactions on 21~(6), 1170--1175.

\bibitem[{Hofmann and Buhmann(1997)}]{hofmann1997pairwise}
Hofmann, T., Buhmann, J.~M., 1997. Pairwise data clustering by deterministic
  annealing. Pattern Analysis and Machine Intelligence, IEEE Transactions on
  19~(1), 1--14.

\bibitem[{Hu et~al.(2008)Hu, Li, Zhang, Fan, and Di}]{hu2008community}
Hu, Y., Li, M., Zhang, P., Fan, Y., Di, Z., 2008. Community detection by
  signaling on complex networks. Physical Review E 78~(1), 016115.

\bibitem[{Islam and Brankovic(2011)}]{islam2011privacy}
Islam, M.~Z., Brankovic, L., 2011. Privacy preserving data mining: a noise
  addition framework using a novel clustering technique. Knowledge-Based
  Systems 24~(8), 1214--1223.

\bibitem[{Jain(2010)}]{jain2010data}
Jain, A.~K., 2010. Data clustering: 50 years beyond k-means. Pattern
  Recognition Letters 31~(8), 651--666.

\bibitem[{Jiang et~al.(2012)Jiang, Jia, and Yu}]{jiang2012efficient}
Jiang, Y., Jia, C., Yu, J., 2012. An efficient community detection method based
  on rank centrality. Physica A: Statistical Mechanics and its Applications.

\bibitem[{Lancichinetti et~al.(2009)Lancichinetti, Fortunato, and
  Kert{\'e}sz}]{lancichinetti2009detecting}
Lancichinetti, A., Fortunato, S., Kert{\'e}sz, J., 2009. Detecting the
  overlapping and hierarchical community structure in complex networks. New
  Journal of Physics 11~(3), 033015.

\bibitem[{Lancichinetti et~al.(2008)Lancichinetti, Fortunato, and
  Radicchi}]{lancichinetti2008benchmark}
Lancichinetti, A., Fortunato, S., Radicchi, F., 2008. Benchmark graphs for
  testing community detection algorithms. Physical Review E 78~(4), 046110.

\bibitem[{Liu and Liu(2010)}]{liu2010detecting}
Liu, J., Liu, T., 2010. Detecting community structure in complex networks using
  simulated annealing with k-means algorithms. Physica A: Statistical Mechanics
  and its Applications 389~(11), 2300--2309.

\bibitem[{Liu et~al.(2013)Liu, Pan, and Dezert}]{liu2013evidential}
Liu, Z.-g., Pan, Q., Dezert, J., 2013. Evidential classifier for imprecise data
  based on belief functions. Knowledge-Based Systems 52, 246--257.

\bibitem[{Martin and Quidu(2008)}]{martin2008decision}
Martin, A., Quidu, I., 2008. Decision support with belief functions theory for
  seabed characterization. In: Information Fusion, 2008 11th International
  Conference on. IEEE, pp. 1--8.

\bibitem[{Masson and Denoeux(2008)}]{masson2008ecm}
Masson, M.-H., Denoeux, T., 2008. {ECM}: An evidential version of the fuzzy
  $c$-means algorithm. Pattern Recognition 41~(4), 1384--1397.

\bibitem[{Masson and Den{\oe}ux(2009)}]{masson2009recm}
Masson, M.-H., Den{\oe}ux, T., 2009. {RECM}: Relational evidential $c$-means
  algorithm. Pattern Recognition Letters 30~(11), 1015--1026.

\bibitem[{Mendes and Sacks(2003)}]{mendes2003evaluating}
Mendes, M., Sacks, L., 2003. Evaluating fuzzy clustering for relevance-based
  information access. In: Fuzzy Systems, 2003. FUZZ'03. The 12th IEEE
  International Conference on. Vol.~1. IEEE, pp. 648--653.

\bibitem[{Newman and Girvan(2004)}]{newman2004finding}
Newman, M.~E., Girvan, M., 2004. Finding and evaluating community structure in
  networks. Physical review E 69~(2), 026113.

\bibitem[{Raghavan et~al.(2007)Raghavan, Albert, and Kumara}]{raghavan2007near}
Raghavan, U.~N., Albert, R., Kumara, S., 2007. Near linear time algorithm to
  detect community structures in large-scale networks. Physical Review E
  76~(3), 036106.

\bibitem[{Sales-Pardo et~al.(2007)Sales-Pardo, Guimera, Moreira, and
  Amaral}]{sales2007extracting}
Sales-Pardo, M., Guimera, R., Moreira, A.~A., Amaral, L. A.~N., 2007.
  Extracting the hierarchical organization of complex systems. Proceedings of
  the National Academy of Sciences 104~(39), 15224--15229.

\bibitem[{Schubert(2004)}]{schubert2004clustering}
Schubert, J., 2004. Clustering belief functions based on attracting and
  conflicting metalevel evidence using potts spin mean field theory.
  Information Fusion 5~(4), 309--318.

\bibitem[{Smets(2005)}]{smets2005decision}
Smets, P., 2005. Decision making in the {TBM}: the necessity of the pignistic
  transformation. International Journal of Approximate Reasoning 38~(2),
  133--147.

\bibitem[{Smyth and White(2005)}]{smyth2005spectral}
Smyth, S., White, S., 2005. A spectral clustering approach to finding
  communities in graphs. In: Proceedings of the 5th SIAM International
  Conference on Data Mining. pp. 76--84.

\bibitem[{Tabassian et~al.(2012)Tabassian, Ghaderi, and
  Ebrahimpour}]{tabassian2012combining}
Tabassian, M., Ghaderi, R., Ebrahimpour, R., 2012. Combining complementary
  information sources in the {D}empster--{S}hafer framework for solving
  classification problems with imperfect labels. Knowledge-Based Systems 27,
  92--102.

\bibitem[{Tasgin et~al.(2007)Tasgin, Herdagdelen, and
  Bingol}]{tasgin2007community}
Tasgin, M., Herdagdelen, A., Bingol, H., 2007. Community detection in complex
  networks using genetic algorithms. arXiv preprint arXiv:0711.0491.

\bibitem[{Yang et~al.(2013)Yang, Di, Liu, and Liu}]{yang2013hierarchical}
Yang, B., Di, J., Liu, J., Liu, D., 2013. Hierarchical community detection with
  applications to real-world network analysis. Data \& Knowledge Engineering
  83, 20--38.

\bibitem[{Yang et~al.(2011)Yang, Zhang, Lu, and Ma}]{yang2011kernel}
Yang, X., Zhang, G., Lu, J., Ma, J., 2011. A kernel fuzzy c-means
  clustering-based fuzzy support vector machine algorithm for classification
  problems with outliers or noises. Fuzzy Systems, IEEE Transactions on 19~(1),
  105--115.

\bibitem[{Zachary(1977)}]{zachary1977information}
Zachary, W.~W., 1977. An information flow model for conflict and fission in
  small groups. Journal of anthropological research, 452--473.

\bibitem[{Zhang et~al.(2007)Zhang, Wang, and Zhang}]{zhang2007identification}
Zhang, S., Wang, R.-S., Zhang, X.-S., 2007. Identification of overlapping
  community structure in complex networks using fuzzy c-means clustering.
  Physica A: Statistical Mechanics and its Applications 374~(1), 483--490.

\bibitem[{Zhang et~al.(2010)Zhang, Yoshida, Tang, and Wang}]{zhang2010text}
Zhang, W., Yoshida, T., Tang, X., Wang, Q., 2010. Text clustering using
  frequent itemsets. Knowledge-Based Systems 23~(5), 379--388.

\bibitem[{Zhang et~al.(2013)Zhang, Zhu, Wang, and Zhao}]{zhang2013identifying}
Zhang, X., Zhu, J., Wang, Q., Zhao, H., 2013. Identifying influential nodes in
  complex networks with community structure. Knowledge-Based Systems 42,
  74--84.

\bibitem[{Zhou(2003)}]{zhou2003distance}
Zhou, H., 2003. Distance, dissimilarity index, and network community structure.
  Physical review {E} 67~(6), 061901.

\bibitem[{Zhou et~al.(2014)Zhou, Martin, and Pan}]{zhou2014evidential}
Zhou, K., Martin, A., Pan, Q., 2014. Evidential communities for complex
  networks. In: Information Processing and Management of Uncertainty in
  Knowledge-Based Systems. Springer, pp. 557--566.

\end{thebibliography}

\end{document}